\documentclass[manuscript,nonacm]{acmart}
\usepackage{hyperref}
\usepackage{color}
\usepackage{xcolor}
\usepackage{amsmath}
\usepackage[graphicx]{realboxes}

\usepackage{amsmath}
\usepackage{booktabs}
\usepackage{multirow}
\usepackage{xspace}

\graphicspath{{figs/}}

\newcommand{\para}[1]{\smallskip\noindent{\bf #1}}
\newcommand{\figref}[1]{Figure~\ref{#1}}
\newcommand{\tbref}[1]{Table~\ref{#1}}
\newcommand{\secref}[1]{\S\ref{#1}}
\newcommand{\shallow}{``shallow''\xspace}
\newcommand{\Shallow}{``Shallow''\xspace}

\begin{document}

\title{Towards a Science of Human-AI Decision Making: A Survey of Empirical Studies
}

\author{Vivian Lai}
\affiliation{%
  \institution{University of Colorado Boulder}
  \country{USA}
}
\author{Chacha Chen}
\affiliation{%
  \institution{University of Chicago}
  \country{USA}
}
\author{Q. Vera Liao}
\affiliation{%
  \institution{Microsoft Research}
  \country{Canada}
}
\author{Alison Smith-Renner}
\affiliation{%
  \institution{Dataminr}
  \country{USA}
}
\author{Chenhao Tan}
\affiliation{%
  \institution{University of Chicago}
  \country{USA}
}

\date{}

\begin{abstract}

As AI systems demonstrate increasingly strong predictive performance, their adoption has grown in numerous domains.  However, in high-stakes domains such as criminal justice and healthcare, full automation is often not desirable due to safety, ethical, and legal concerns, yet fully manual approaches can be inaccurate and time consuming. As a result, there is growing interest in the research community to augment human decision making with AI assistance. Besides developing AI technologies for this purpose, the emerging field of human-AI decision making must embrace empirical approaches to form a foundational understanding of how humans interact and work with AI to make decisions.  To invite and help structure research efforts towards a science of understanding and improving human-AI decision making, we survey recent literature of empirical human-subject studies on this topic. We summarize the study design choices made in over 100 papers in three important aspects:  (1) decision tasks, (2) AI models and AI assistance elements, and (3) evaluation metrics.
For each aspect, we summarize current trends, discuss gaps in current practices of the field, and make a list of recommendations for future research.
Our survey highlights the need to develop common frameworks to account for the design and research spaces of human-AI decision making, so that researchers can make rigorous choices in study design, and the research community can build on each other's work and produce generalizable scientific knowledge. We also hope this survey will serve as a bridge for HCI and AI communities to work together to mutually shape the empirical science and computational technologies for human-AI decision making.

\end{abstract}

\maketitle

\section{Introduction}
\label{sec:intro}

\begin{figure}[t]
    \centering
    \includegraphics[width=.75\textwidth]{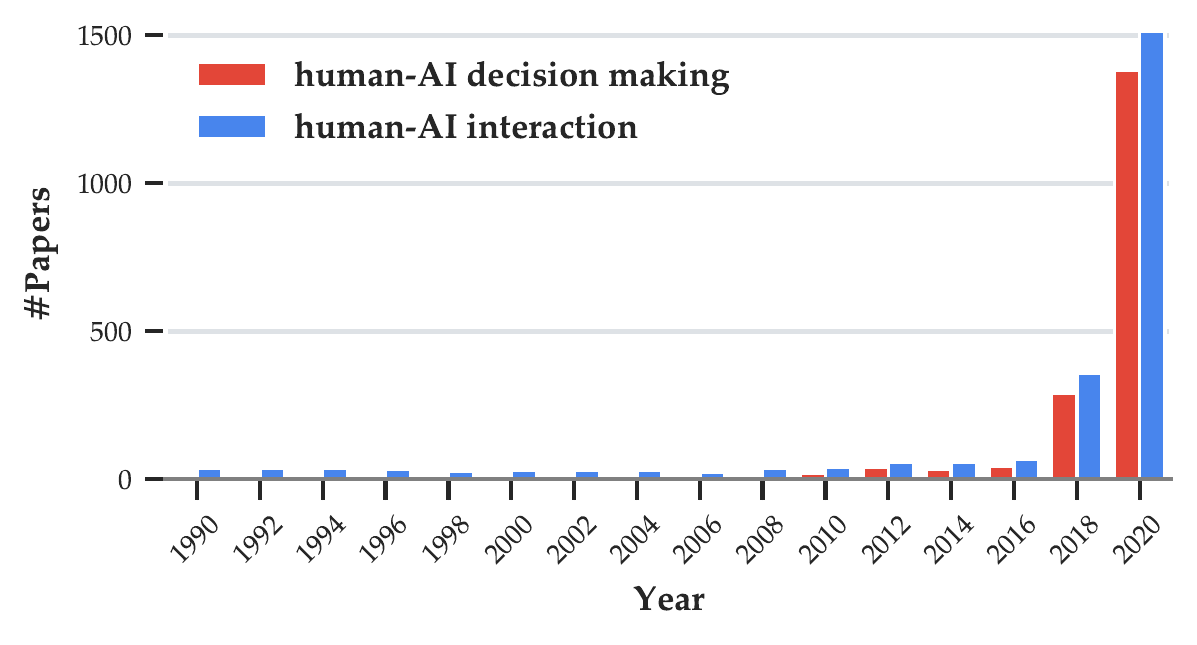}
    \caption{The number of papers based on Google Scholar for two queries, human-AI interaction and human-AI decision making, over the past years.}
    \label{fig:keywords_paper}
\end{figure}

Thanks to recent advances, AI has become a ubiquitous technology and been introduced into high-stakes domains such as healthcare, finance, criminal justice, hiring, and more \cite{annappindi2014system,khandani2010consumer,yang2018insurance,reuters2018,dilsizian2014artificial,wang2014improving}. To prevent hazardous consequences of failures, complete automation is often not desirable in such domains. Instead, AI systems are often introduced to augment or assist human decision makers---by providing a prediction or recommendation for a decision task, which humans can choose to follow or ignore and make their own decision. Besides predictions, current AI technologies can provide a range of other capabilities to help humans gauge the model predictions and make better final decisions, such as providing performance metrics of the model, or uncertainty and explanation for the prediction. In this paper, we refer to these capabilities as different \textit{AI assistance elements} that an AI system can choose to provide. We refer to this general paradigm as human-AI decision making, though in relevant literature we found a multitude of generic terms used such as human-AI interaction, human-AI collaboration, and human-AI teaming. \figref{fig:keywords_paper} shows that the number of papers on this topic has been growing dramatically in the past five years. 

Many of these papers made technical contributions for AI to better support human decisions, such as developing new algorithms to generate explanations for model predictions. 
Meanwhile, the research community is increasingly recognizing the importance of empirical studies of human-AI decision making by involving human subjects and performing decision tasks. 
These studies are not only necessary to \textit{evaluate} the effectiveness of AI technologies in assisting decision making, but also to form a \textit{foundational understanding} of how people interact with AI to make decisions. 
This understanding can serve multiple grounds, including: (1) to inform new AI techniques that provide more effective assistance or are more human compatible; (2) to guide practitioners in making technical and design choices for more effective decision-support AI systems; (3) to provide input for human-centric policy, infrastructure, and practices around AI for decision making, such as regulatory requirements on when AI can or cannot be used to augment certain human decisions~\citep{wicourts}.

However, empirical human-subject studies on human-AI decision making are distributed in multiple research communities, asking diverse research questions, and adopting various methodologies. Currently there is a lack of coherent overview of this area, let alone coherent practices in designing and conducting these studies, hindering concerted research effort and emergence of scientific knowledge. We recognize several challenges to coherence. First, empirical studies of human-AI decision making are conducted in various domains with different decision tasks. Without investigating the scope of these tasks and their impact,  we may not be able to generalize from individual findings. Second, human interactions with AI are enabled and augmented by the affordances of chosen AI assistance elements. Individual empirical studies tend to focus on a small set of AI assistance elements. There is a lack of common frameworks to understand how results for these AI assistance elements generalize and, therefore, their effect on human interactions with AI. Lastly, design of human-subject studies are inherently complex, varying depending on the research questions, disciplinary practices, accessible subjects and other resources, etc. This challenge is further exacerbated by the fact that methodologies, including evaluation metrics, to study human-AI interaction are still in an early development stage.

To facilitate coherence and develop a rigorous science of human-AI decision making, we provide an overview of the current state of the field --- focusing on \textit{empirical human-subject studies} of human-AI decision making --- through this survey. 
We focus on studies with the goal of evaluating, understanding and/or improving human performance and experience for a decision making task, rather than improving the model.
This scope differentiates from prior surveys on empirical studies of human-AI interactions that either deviate from the scope of decision making or focus on only one aspect of it, like trust~\cite{sperrle2021survey, sperrle2020should, hohman2018visual, wu2021survey, lertvittayakumjorn2021explanation, vereschak2021evaluate}. 
With the above-mentioned challenges in mind,
our survey focuses on analyzing three aspects of study design choices made in these surveyed papers: the decision tasks, the types of AI and AI assistance elements, and the evaluation metrics. For each aspect we summarize current trends, identify potential gaps, and provide recommendations for future work.

The remainder of this survey paper is structured as follows. 
We first discuss the scope of papers surveyed, and methodology for paper inclusion and coding. 
We then present our analysis for each of the three areas mentioned above, with summary tables provided. 
We conclude the survey with a call to actions for developing common frameworks for the design and research spaces of human-AI decision making, and mutual shaping of empirical science and computational technologies in this emerging area.  
To allow for easy access to the papers that we cite, they are available at \url{https://haidecisionmaking.github.io}.

\section{Methodology}
\label{sec:method}

In this section, we define the scope of our survey in detail and describe how we selected papers to include.

\para{Survey scope and paper inclusion criteria.}
The focus of this survey is on \textit{empirical human-subject studies} of \textit{human-AI decision making}, where the goal is to evaluate, understand and/or improve human performance and experience for a decision making task, rather than to improve the model. As such, we specify the following inclusion criteria: 

\begin{itemize}
	\item The paper must include evaluative human-subject studies. We thus exclude purely formative studies that focus on exploring user needs to inform design of AI systems, often qualitatively.
	\item The paper must target a decision making task, thus we exclude tasks of other purposes (e.g., debugging and other forms of improving the model, co-creation, gaming).
	\item The task must involve and focus on studying human decision makers, thus we exclude papers on AI automation or other AI stakeholders (model developers, ML practitioners). However, we do not limit our studies to those that implement complete decision making processes, but also include studies that claim to evaluate some aspects of decision makers' perceptions, such as their understanding, satisfaction, and perceived fairness of the AI.

\end{itemize}

\para{Search strategy.}
In addition to papers that we were already aware of, we looked through proceedings of premier conferences where AI and human-computer interaction (HCI) works are published from 2018 to 2021, to identify papers that fit the criteria mentioned above.
Specifically, conferences we searched include
ACM CHI Conference on Human Factors in Computing Systems,
ACM Conference on Computer-supported Cooperative Work and Social Computing,
ACM Conference on Fairness, Accountability, and Transparency,
ACM Conference on Intelligent User Interfaces,
Conference on Empirical Methods in Natural Language Processing,
Conference of the Association for Computational Linguistics, and 
Conference of the North American Chapter of the Association for Computational Linguistics.
We focused on NLP conferences in AI because (1) the fraction of papers with empirical studies on human subjects is low in AI (including NLP) conferences such as AAAI and NeurIPS; (2) the expertise of the authors enables regular examinations of all papers in NLP conferences.
We further expand our considered papers by choosing relevant references from each paper and examining the papers that cite these papers. 

An initial search process yielded in 
over 130 papers.
All the authors then looked through the list and discussed with each other to exclude out-of-scope papers, resulting in
over 80
in-scope papers. We collated them in a spreadsheet then started coding these papers by the three study design choices they make: decision tasks, types of AI and AI assistance elements, and evaluation metrics.

We started the coding by having one author extracting relevant information from the paper, such as what kind of decision tasks, AI models and assistance elements, and evaluation metrics were used in the study. A second-round coding was then performed, focusing on merging similar codes and grouping related codes into areas. For example, decision tasks were grouped by domains, as shown in Table~\ref{sec:tasks}.
For the second-round coding, all authors were assigned to look through the initial codes for one of the three study design choices. All authors met regularly to discuss the grouping and ambiguous cases to ensure the integrity of the process, until consensus on the codes and grouping, as presented in the summary tables, were reached.
{\em We provide summary tables for each study design choice and these tables can provide a quick overview of the literature space.}

\section{Decision Tasks}
\label{sec:tasks}
We start by reviewing decision tasks that researchers have used to conduct empirical studies of human-AI decision making.
We group decision tasks used in prior studies based on their application domains (e.g., healthcare, education, etc.).
To facilitate consideration on how the choices of decision tasks may impact the generalizability of study results, we highlight four dimensions that differ in these tasks---risk, required expertise, subjectivity, and AI for emulation vs. discovery---to help interpret results in these studies.

\subsection{Tasks Grouped by Domains}

We group the decision tasks used in the surveyed papers by their application domains, as summarized in \tbref{tb:tasks}: Law \& Civic, Medicine \& Healthcare, Finance \& Business, Education, Leisure, Professional, Artificial, Generic, and others.

\para{Law \& Civic}. 
This domain includes tasks in the justice system and for civil purposes.
The most commonly used task is recidivism prediction,
which has attracted a lot of interest since the ProPublica article on biases in recidivism prediction algorithms used by the criminal justice system in the United States~\cite{angwin2016machine}.
Popular datasets for recidivism prediction and the discussed variants include COMPAS \citep{angwin2016machine}, rcdv \cite{schmidt1988predicting}, and ICPSR \citep{icpsr}.
Given these datasets, how these studies defined recidivism prediction varied.
A common formulation is to predict whether a person with a particular profile will recidivate, or be rearrested within 2 years of their most recent crime~\citep{liu2021understanding,wang2021explanations,van2021effect} or reoffend before bail \citep{grgic2019human}. 
Slight variations of this definition include: (1) predicting if a rearrest is for a {\em violent} crime within two years \citep{green2019principles,green2019disparate} or
(2) predicting if the defendant will reoffend or not, without the two year time limit \citep{dodge2019explaining,ribeiro2018anchors}.
In comparison, \citet{green2019principles,green2019disparate}
define the recidivism prediction task not as a binary prediction, but as assessing the likelihood that the defendant 
will commit a crime or fail to appear in court if they are released before trial, ranging from a scale from 0\% to 100\% in intervals of 10\%.
This more fined-grained question could potentially make it harder for human subjects to assess and use model output.  
In a slightly different set-up, \citet{lakkaraju2016interpretable,harrison2020empirical,anik2021data} define the decision tasks as either to
predict if
a defendant is released on bail
or predict one out of four given bail outcomes: (1) the defendant is not arrested while out on
bail and appears for further court dates (No Risk); (2) the defendant fails to appear for further court dates (FTA); (3) the defendant commits
a non-violent crime (NCA) and (4) the defendant commits a violent crime (NVCA) when released on bail.

\begin{table}[htbp]
\centering

\begin{tabular}{p{0.25\textwidth}|p{0.75\textwidth}}
\toprule
Domain & Decision task \\ 

\midrule
Law \& Civic & 
Recidivism prediction 
\citep{liu2021understanding,wang2021explanations,van2021effect,grgic2019human}
and its slight variations \citep{dodge2019explaining,ribeiro2018anchors},
likelihood to recidivate \citep{green2019principles,green2019disparate,liu2021understanding},
bail outcomes prediction \citep{lakkaraju2016interpretable,harrison2020empirical,anik2021data},
child maltreatment risk prediction \citep{de2020case}.
\\

\midrule
Medicine \& Healthcare & 
Medical disease diagnosis \citep{lakkaraju2016interpretable},
cancer image search \citep{cai2019human},
cancer image classification \citep{kiani2020impact},
COVID-19 diagnosis \citep{tsai2021exploring},
balance disorder diagnosis \citep{bussone2015role},
clinical notes annotation/medical coding \citep{levy2021assessing},
stroke rehabilitation assessment \citep{lee2020co,lee2021human}
\\

\midrule
Finance \& Business & 
Income prediction \citep{ribeiro2018anchors,weerts2019human,zhang2020effect,hase2020evaluating},
loan approval~\citep{green2019principles,binns2018s,van2021effect},
loan risk prediction \citep{chromik2021think},
sales forecast \citep{dietvorst2015algorithm,lucic2020does},
property price prediction \citep{poursabzi2018manipulating,abdul2020cogam},
apartment price prediction~\cite{mcgrath2020does}
selecting overbooked airline passengers for re-routing \citep{binns2018s},
determining to freeze bank accounts due to money laundering suspicion \citep{binns2018s},
stock price prediction \citep{biran2017human},
marketing email prediction \citep{guo2019visualizing}, 
dynamically pricing car insurance premiums \citep{binns2018s}.
\\

\midrule
Education & 
Students' performance forecasting \citep{dietvorst2015algorithm,dietvorst2018overcoming,weerts2019human},
student admission prediction \citep{anik2021data,cheng2019explaining},
student dropout prediction \citep{lakkaraju2016interpretable}, 
LSAT question answering \citep{bansal2020does}.
\\

\midrule
Leisure & 
Music recommendation \citep{kulesza2013too,kulesza2012tell},
movie recommendation \citep{kunkel2019let},
song rank order prediction \citep{logg2019algorithm},
speed dating \citep{yin2019understanding,lu2021human},
Facebook news feed prioritization \citep{rader2018explanations},
Quizbowl \citep{feng2018can},
draw-and-guess \citep{cai2019effects}
word guessing \citep{gero2020mental},
chess playing \citep{das2020leveraging},
plant classification \citep{yang2020visual},
goods division \citep{lee2019procedural}.
\\

\midrule
Professional & 
Job promotion \citep{binns2018s},
meeting scheduling assistance \citep{kocielnik2019will},
email topic classification \citep{smithrenner2020},
cybersecurity monitoring \citep{ehrlich2011taking},
profession prediction \citep{liu2021understanding}, 
military planning (monitor and direct unmanned vehicles) \citep{stowers2017insights}. 
\\

\midrule
Artificial & 
Alien medicine recommendation \citep{lage2019evaluation,narayanan2018humans},
alien recipe recommendation \citep{lage2019evaluation,narayanan2018humans},
jellybean counting \citep{park2019slow},
broken glass prediction \citep{yu2019trust},
defective object pipeline \citep{bansal2019beyond,bansal2019updates}, 
news reading time prediction \citep{szymanski2021visual},
water pipe failure prediction \citep{arshad2015investigating},
math questions \citep{friedler2019assessing}.
\\

\midrule
Generic & 
Question answering \citep{chandrasekaran2018explanations,gonzalez2020human},
image classification \citep{alqaraawi2020evaluating},
review sentiment analysis \citep{nguyen2018comparing,bansal2020does,hase2020evaluating}.
\\

\midrule
Others & 
Deception detection \citep{lai2020chicago,lai2019human,liu2021understanding},
forest cover prediction \citep{wang2021explanations},
toxicity classification \citep{carton2020feature},
nutrition prediction \citep{buccinca2020proxy,buccinca2021trust},
person weight estimation \citep{logg2019algorithm},
attractiveness estimation \citep{logg2019algorithm},
activity recognition \citep{lim2009and,nourani2021anchoring},
emotion analysis \citep{springer2018progressive},
religion prediction \citep{ribeiro2016should}.

\\

\bottomrule
\end{tabular}

\caption{Types of decision tasks grouped by application domains in human-AI decision making.
}
\label{tb:tasks}
\end{table}

Prior works also
explore how AI assistance can be applied to civil activities in the public sector.
For example, \citet{de2020case} examine the child maltreatment hotline 
and develop a model that assists call workers in identifying potential high-risk cases.

\para{Medicine \& Healthcare.}
This domain includes tasks related to clinical decision making,
ranging from medical diagnosis to sub tasks such as medical image search. 
The general formulation of medical diagnosis is to predict whether a patient has a disease given a list of symptoms or other information about the patient.
Researchers have studied AI assistance for a range of medical diagnosis tasks, include general disease diagnosis \citep{lakkaraju2016interpretable}, COVID-19 diagnosis \citep{tsai2021exploring} and balance disorder diagnosis \citep{bussone2015role}. 
Another popular area is imaging-related assistance to help medical staff 
make better decisions during medical diagnoses.
For example, \citet{cai2019human} developed a tool for pathologists to search for similar images when diagnosing prostate cancer.
Other imaging tasks include interpreting chest x-rays \cite{kiani2020impact}. 
In addition, \citet{lee2020co,lee2021human} investigate a support system for stroke rehabilitation assessment, which assists physical therapists in assessing patients' progress.
Due to the difficulty in understanding and predicting biological processes in healthcare, these tasks can be difficult, even for medical experts (i.e., pathologists, radiologists).
Lastly, unlike previously mentioned studies where the focus is on medical diagnosis, \citet{levy2021assessing} investigates the effects on annotating medical notes with the help of AI assistance.

\para{Finance \& Business.}
This domain includes decisions related to income, businesses, and properties.
Popular datasets used in income and credit prediction include the Adult Income dataset in UCI Machine Learning Repository \cite{adult} and Lending Club \cite{lending}.
As a result, an income task is to predict whether a particular person's profile earns more than \$50K annually \citep{ribeiro2018anchors,weerts2019human,zhang2020effect,hase2020evaluating}, driven by the Adult Income dataset \cite{adult}.
Note that the number, \$50K, is outdated and somewhat arbitrary given inflation. 
Other similar tasks include loan approval (e.g., assessing the likelihood of an applicant defaulting on a loan) \citep{green2019principles,binns2018s,van2021effect}, loan risk prediction \citep{chromik2021think}, and freezing of bank account prediction \citep{binns2018s}.

In addition to income and credit prediction in the financial domain, prior work that explores how AI assistance can help make other business-related decisions.
Some classification tasks include sales forecasting prediction where \citet{dietvorst2015algorithm} define a sales forecasting task to predict the rank (1 to 50) of individual U.S. states in terms of the number of airline passengers that departed from that state in 2011,
marketing email prediction \citep{guo2019visualizing} where the task is to predict the better email to send given customers reactions,
and predicting overbooked airline flights \citep{binns2018s},
On the other hand, regression tasks include forecasting monthly sales of Ahold Delhaize's stores \cite{lucic2020does},
property price prediction \citep{poursabzi2018manipulating,abdul2020cogam},
apartment rent prediction \citep{mcgrath2020does},
and car insurance prediction \citep{binns2018s}.
Lastly, \citet{biran2017human} asks participants to decide if they would buy a stock given various AI assistance.

\para{Education.}
This domain includes decisions performed within the education system. 
Most tasks are broadly about forecasting student performance.
Different variations exist: (1) predicting how well students perform in a given program \cite{dietvorst2015algorithm};  (2) predicting if a student will not graduate on time or drop out \citep{lakkaraju2016interpretable}
(3) predicting students' grades in tests such as math exams \cite{dietvorst2018overcoming,weerts2019human}; 
and (4) making admission decisions \citep{anik2021data,cheng2019explaining}.
\citet{bansal2020does} also considered the test questions in Law School Admission Test (LSAT).

\para{Leisure.}
This domain includes tasks serving entertainment purposes. 
Prior works have explored AI assistance for a range of leisure activities  (e.g., 
to recommend music \citep{kulesza2013too,kulesza2012tell},
to recommend movies \citep{kunkel2019let},
to predict songs' chart ranking (i.e. song popularity) \citep{logg2019algorithm}),
and to reorder your Facebook news feed \citep{rader2018explanations}. 
Other works used games or gamified tasks such as to predict if a person would date a person given a profile \citep{yin2019understanding,lu2021human},
classify types of leaves \citep{yang2020visual},
and distribute goods fairly \citep{lee2019procedural}.
Games are also used, such as Quizbowl \cite{feng2018can},
draw-and-guess~\cite{cai2019effects}, word-guessing~\cite{gero2020mental}, and chess playing \citep{das2020leveraging}.

\para{Professional.}
This domain includes tasks related to employment and professional progress.
\citet{binns2018s} define a task to predict whether a person's profile would receive a promotion.
\citet{liu2021understanding} define a task to predict a person's occupation given a short biography.
Other tasks not related to jobs include classifying emails' topic \citep{smithrenner2020},
AI-assisted meeting scheduling \citep{kocielnik2019will},
military planning via monitoring and unmanned vehicles \citep{stowers2017insights},
and cybersecurity monitoring \citep{ehrlich2011taking}.

\para{Artificial.}
This domain includes tasks that are artificial or fictional, usually made up to explore specific research questions.
\citet{lage2019evaluation} and \citet{narayanan2018humans} created two fictional tasks to evaluate the effect of providing model explanations:
(1) predicting aliens food preferences in various settings and (2) personalized treatment strategies for various fictional symptoms. 
Other tasks include predicting the number of jellybeans in an image \citep{park2019slow},
predicting water pipe failure \citep{arshad2015investigating},
predicting news reading time \citep{szymanski2021visual},
predicting broken glass \citep{yu2019trust},
predicting defective object in a pipeline \citep{bansal2019beyond,bansal2019updates},
and answering math questions \citep{friedler2019assessing}.
Artificial tasks have the advantage of being easily accessible to lay people, and allowing researchers to control for confounding factors.
The flip side is that results obtained from artificial tasks may not generalize to real applications.

\para{Generic.}
Generic tasks are ones without specified applications and can be applied to different domains. 
These include AI benchmarks where crowdsourced datasets are used to test how well AI models can emulate human intelligence such as object recognition in images (e.g., horses, trains) \citep{alqaraawi2020evaluating} and \citep{chandrasekaran2018explanations,gonzalez2020human}.
Another popular generic task is review sentiment analysis, which are performed with various contents such as movie reviews \citep{nguyen2018comparing,hase2020evaluating}, beer reviews \citep{bansal2020does},
and book reviews \citep{bansal2020does}.

\para{Others.}
Finally, we list decision tasks that do not fit in any of the domains above:
attractiveness estimation~\cite{logg2019algorithm};
activity recognition (e.g., exercise \citep{lim2009and} and kitchen~\cite{nourani2021anchoring});
deception detection, predicting if a hotel review is deceptive \citep{lai2020chicago,lai2019human,liu2021understanding};
toxicity classification \citep{carton2020feature}, predicting external consensus of whether a comment is toxic or not;
predicting a person's weight based on an image \citep{logg2019algorithm}; predicting the nutrition value of a dish given an image \citep{buccinca2020proxy,buccinca2021trust};
religion prediction \citep{ribeiro2016should}, predicting whether text is about Christianity or atheism;
emotion analysis \citep{springer2018progressive}, predicting emotion of text;
forest cover prediction, predicting if an area is covered by spruce-fir forest \citep{wang2021explanations}.

\subsection{Task Characteristics}
\label{task-char}

Given the wide variety of decision tasks that have been studied, it is important to understand how findings generalize across tasks.
Although domain can serve as a thematic umbrella, it is not useful for evaluating generalizability because each domain includes tasks with drastically different properties (e.g., medicine \& healthcare includes both diagnosing cancer and annotating medical notes).
Here we seek to identify meaningful task characteristics.

Characteristics of the decision task can determine whether a task is appropriate for the claims in a study as well as their generalizability. For example,  a low-stakes decision task may not create an ideal condition with vulnerability to study trust. A task that is more challenging for human to perform may induce higher baseline reliance on the AI, so the results may not generalize to settings where the human outperforms the AI, and vice versa. However, existing literature often does not provide explicit justification on the choices of decision task, nor indicate the scope of generalizability of the results.

To facilitate future research to make such considerations, we look across the surveyed papers and highlight
four \textit{dimensions} that vary in the chosen design tasks: (1) task risk (e.g., high, low, or artificial stakes),  (2) required expertise in the task, and (3) decision subjectivity, and (4) AI for emulation vs. discovery. 
We do not claim these four as an exhaustive list, but hope to illuminate the challenges in interpreting and generalizing from results in studies that adopt different decision tasks, and encourage future studies to justify the choice of tasks and report their characteristics.

\para{Risk.}
The risk, including its stakes and potential societal impact, of a task (whether high, low, or artificial) is an important characteristic that could impact decision behaviors, particularly for user trust and reliance.
In fact, \citet{jacovi2021formalizing} argue that trust can only be manipulated and evaluated in high-stakes scenarios where vulnerability is at play.

Tasks in Law \& Civic, Medicine \& Healthcare, Education and Finance \& Business, are mostly considered high stakes.
In comparison, leisure and artificial are mostly of relatively low stakes.
Tasks in Professional can have varying stakes, for example, human resource related decisions are high stakes, while email topic classification is low stakes.
Generic is a category driven by the creation of AI benchmark datasets.
It is unclear how to interpret their stakes or societal relevance, and their stakes are contingent on the contexts they are adopted.

Researchers should carefully consider design choices with respect to risk: it is also critical to be cognizant of potential ethical concerns when using AI assistance for high-stakes decisions, such as recidivism prediction~\cite{green2021flaws}. 
Increasing the unwarranted trust can be highly problematic in high-stakes decisions~\citep{jacovi2021formalizing}.
Moreover, generalizability may be affected by risk as well; for example, findings related to AI assistance effectiveness in the context of low-risk scenarios, especially on reliance, may not generalize to high-risk scenarios without further research. 

\para{Required expertise.}
Levels of expertise or prior training in a task can lead to different decision behaviors with AI. 
For some tasks, limited to no domain expertise is required (e.g., artificial tasks), whereas others require significant expertise (e.g., cancer image classification). 
In the ideal situation, participants in user studies for 
tasks that require domain expertise
should be experts (i.e., judges, radiologists) or target users, however due to pragmatic reasons, some of the studies are performed by laypeople on crowdsourcing platforms.
For instance, it is unclear whether results derived from crowdworkers would generalize to judges or doctors.

Expertise required is often correlated with risk.
High-risk tasks in Law \& Civic, Medicine \& Healthcare, Education and Finance \& Business usually require expertise, while low-risk tasks in leisure and artificial do not.
Similarly, tasks in Professional can have varying requirements in expertise, for example, human resource related decisions may require more expertise than  email topic classification.
Generic usually do not require expertise.

While many works in human-AI decision making categorize decision makers as either ``domain experts'' or ``lay users'', AI literacy is also an important consideration, especially as it relates to one's ability to interpret AI assistance elements.
A framework proposed by~\citet{suresh2021beyond} suggests decomposing stakeholder expertise into both context (domain vs. machine learning) and knowledge for that context (e.g., formal, personal, etc.).
For example, decision making performance with AI-enabled medical decision support tools~\cite{cai2019human}, is affected by both formal, instrumental, and personal domain expertise as well as instrumental machine learning expertise (i.e., familiarity with ML toolkits). 
As such, systems should be evaluated with targeted expertise, or even varied levels of expertise to investigate the generalizability of results. Studies should also carefully report on participants' expertise to allow appropriate interpretation and usage of the results. 

\para{Subjectivity.}
Many decision tasks are framed as supervised prediction problems in machine learning, where there exists groundtruth, $y$.
This choice often implicitly assumes that this is an objective prediction task (at least in hindsight), e.g., whether a person has a balance disorder or not~\cite{bussone2015role} or whether a person will pay back a loan~\cite{binns2018s,green2019principles}.
Only in these tasks, quantitative measures of human performance are appropriate.
In comparison, personal decision making can be subjective, for example, whether a music recommendation is good is subjective for the person receiving it~\cite{kulesza2013too,kulesza2012tell}; similarly, whether or not language is perceived as ``toxic'' depends on the person assessing it and their determination is hard for others to refute~\cite{carton2020feature}. 
Subjective decision tasks typically have high variability (low agreement) on what is the \textit{correct} model output. Yet, AI assistance is still valuable to help people make subjective decisions. 
However, human performance might not be a good measure for evaluating these subjective decision making tasks, or non-trivial assumptions are required to convert such decisions to objective tasks (e.g., predicting which movie has the highest box office proceeds). 
As a major focus so far in human-AI decision making is to improve the performance of human-AI teams, most of the tasks in our surveyed papers are objective tasks.

\para{AI for emulation vs. discovery.}
We highlight a final dimension that affects how one should interpret results from a study but is often overlooked in the choice of tasks.
Within objective tasks, we can further distinguish tasks based on the source of groundtruth.
In many high-stakes decisions, groundtruth can come from (social and biological) processes that are external to human judgments (e.g., the labels in recidivism prediction are based on observing the behavior of the defendants after bailing rather than judges' decisions).
In these tasks, machine learning models can be used to 
{\em discover} patterns that humans may not recognize, and can be useful for tasks such as  
recidivism prediction \citep{liu2021understanding,wang2021explanations,van2021effect,grgic2019human,dodge2019explaining,ribeiro2018anchors,green2019principles,green2019disparate}, 
deception detection \cite{lai2019human,lai2020chicago,liu2021understanding},
and income prediction \citep{ribeiro2018anchors,weerts2019human,zhang2020effect,hase2020evaluating,van2021effect}.
We refer to such tasks as {\em AI for discovery} tasks.\footnote{Expertise can be orthogonal from discovery/emulation. For example, interpreting a test result based on existing standards is an emulation task (e.g., determining whether one's heart rate is abnormal) for doctors, while deciding the best treatment for a patient with cancer or determining prostate cancer from MRI images is a discovery task because doctors do not have access to the groundtruth and the groundtruth is related to complex biological and social processes beyond the control of doctors.}
These tasks are usually more challenging to humans, because they require humans to reason about 
external (social and biological) processes that are not innate to human intelligence.
In fact, human performance in some of these tasks, such as deception detection tasks~\citep{lai2019human,lai2020chicago}, were found to be close to random guessing (manual annotation is thus {\em inappropriate} for getting groundtruth). Humans decisions are also prone to biases in these challenging reasoning tasks such as recidivism prediction. AI can improve decision efficacy and alleviate potential biases not only by providing predictions, but also by elucidating the embedded patterns in these decision tasks, such as by providing explanations.  However, a challenge lies in the difficulty for humans to determine whether counterintuitive or inconspicuous patterns are genuinely valid or are driven by spurious correlations.

In comparison, a typical narrative of AI is to emulate human intelligence. 
For example, humans perform well at simple recognition tasks, such as determining whether images include people or whether documents discuss sports, and we build AI to emulate this ability.
That is, machine learning models are designed to \textit{emulate} the 
human intelligence for these tasks, and human performance is considered as the upper bound.
In these tasks, the groundtruth comes from human decision makers.
We refer to these tasks as {\em emulation} tasks.
As such, these tasks are designed for automation purposes, and are not preferred choices for studying human-AI decision making because humans are less likely to benefit from AI assistance.
However, there are still a handful of experimental studies investigating human-AI decision making in emulation tasks~\citep{chandrasekaran2018explanations,gonzalez2020human,alqaraawi2020evaluating}.
These tasks are typically in the generic domain, and 
improving human performance might be interpreted as reducing the mistakes of crowdworkers (possibly due to the lack of attention).
It is unclear whether results would generalize to discovery tasks where humans reason about external processes and models may identify counterintuitive patterns, and future research should explicitly consider the boundary between the two.

\subsection{Summary \& Takeaways}

We summarize current trends in the choices of decision tasks, discuss gaps we see in current practices of the field, and make recommendations towards a more rigorous science of human-AI decision making. We follow this organization when summarizing each of the remaining sections. 

\paragraph{Current trends.}

\begin{enumerate}

    \item \textbf{Variety:} Existing studies on human-AI decision making cover a wide variety of tasks in many application domains.
    This variety demonstrates the potential of human-AI decision making and also leads to challenges in generalization of results and developing scientific knowledge across studies.

    \item \textbf{Task characteristics:} Most existing studies focus on high-stake domains such as justice systems, medicine \& healthcare, finance, and education; while artificial and generic tasks are still used by some. Although many decision tasks require domain expertise, experts are seldom the subjects of study.
    Finally, most existing studies focus on ``AI for discovery'' tasks because humans typically need or can benefit from AI's assistance in these tasks more than ``AI for emulation'' tasks. However, studies often do not explicitly justify using decision tasks with these characteristics nor discuss their implications for other study design choices (e.g., subjects) and generalizability of results.

\end{enumerate}

\paragraph{Gaps in current practices.}

\begin{enumerate}

    \item  \textbf{Choice of tasks are driven by datasets availability.}
    For instance, due to the popularity of COMPAS \cite{angwin2016machine} and ICPSR \cite{icpsr} datasets, many studies used recidivism predictions as the decision task and focused on the law \& civic domain.
    In comparison, despite the public discourse on the potential harm of AI in other domains like hiring \cite{reuters2018},
    there is relatively little research on AI assistance in human resources due to lack of available datasets.
    We suspect that this is also the reason that emulation tasks are used in some studies (e.g., prevalence of AI benchmarks such as visual question answering).

    \item \textbf{Lack of frameworks for generalizable knowledge.} 
    A key question for the research community is how to develop scientific knowledge by validating and comparing results from studies across many different domains and types of decision task. For example, when an artificial task is used, how much can the results generalize to other domains?  How to interpret differences in the results in a medical diagnosis task versus a movie recommendation task?
    Do results on medical diagnosis generalize to bailing decisions? 
    We believe a first step is to identify different underlying characteristics of decision tasks such as risk and required expertise, in order to make meaningful comparisons across studies and reconcile differences in empirical results.

    \item \textbf{Misalignment with application reality.} The focuses and study design choices of current studies may not align with how AI is or will be used in real-world decision-support applications. For instance, the overwhelming focus on high-stake domains is worrisome if the study designs (subjects, consequence, context) do not align with the reality.
    Tasks defined based on easily available datasets may deviate from realistic decision making scenarios. For example, experiments based on generic tasks such as visual question answering can be quite different from real-world imaging related tasks such as for medical diagnosis.
    This misalignment is analogous to the discrepancies between the recent burst of COVID-related machine learning papers and clinical practices~\citep{roberts2021common}.

\end{enumerate}   

\paragraph{Recommendations for future work.} 

\begin{enumerate}
    \item \textbf{Develop frameworks to characterize decision tasks.}
    To allow  scientific understanding across studies, there is an urgent need for the field to have frameworks that can characterize the space of human-AI decision tasks. As a starting point we suggest the following dimensions in Section~\ref{task-char}: risk, required expertise, subjectivity and AI for emulation v.s. discovery. We encourage future work to further develop such frameworks. We further encourage specification, such as including meta-data of task characteristics whenever a new decision task or dataset is introduced to study human-AI decision making.

    \item \textbf{Justify choices of decision task.} 
 We encourage researchers to articulate the rationale behind the choice of decision task, including its suitability to answer the research questions. Researchers should also consider whether other study design choices such as system design, subject recruitment and evaluation methods align with the characteristics of the task. Such practices can help interpret and consolidate results across studies and identify important and new dimensions of decision task characteristics.

    \item \textbf{Expand datasets availability.} A bottleneck hindering the community from studying broader and more realistic decision tasks is the availability of datasets. Popular datasets are often introduced for AI algorithmic research and may not reflect what is needed for realistic AI decision-support tasks. The field should motivate dataset creation by what decision tasks are needed to better understand human-AI decision making, which may require first better understanding decision-makers' needs for AI support.
\end{enumerate}

\section{AI Models and AI Assistance Elements}
\label{sec:explanations}
To use AI to accomplish decision tasks, people not only rely on the model's predictions, but can also leverage other information provided by the system to make informed decisions, including gauging if the model predictions are reliable. 
For example, with the recent surge of the field explainable AI (XAI), many have contended that AI explanations could provide additional insights to assist decision making~\cite{lai2019human,lipton2016mythos,doshi2017towards}. 
Therefore, we take a broad view on ``AI assistance elements'' and review the system features studied in prior work that could impact people's decision making. 
 We start with an overview of the types of model and data used in the surveyed studies and then unpack AI assistance elements.

\subsection{AI Models and Data}
An important driving factor for the recent surge of interest in human-AI decision making is the growing capacity of AI models to aid decisions.
This subsection provides a summary of the different types of models used in surveyed studies, as listed in Table~\ref{tb:models}.

\begin{table}[t]
    \centering
    \begin{tabular}{lp{0.7\textwidth}}
    \toprule
    Model & Examples \\ 

    \midrule
    Deep learning models &
    Convolution Neural Networks~\cite{chandrasekaran2018explanations,alqaraawi2020evaluating}, Recurrent Neural Networks~\cite{guo2019visualizing,carton2020feature}, 
    BERT \cite{lai2020chicago}, RoBERTa~\cite{bansal2020does}, 
     VQA model (hybrid LSTM and CNN)~\cite{ribeiro2018anchors},
    not specified~\cite{gero2020mental,gonzalez2020human,hase2020evaluating,kiani2020impact,lee2021human,nourani2021anchoring,cai2019effects,yin2019understanding,cai2019human,peng2019you,ghazvininejad2017hafez,lee2020co,clark2018creative} 
    \\
    
    \midrule
    \Shallow models &  
Logistic regression~\cite{friedler2019assessing,lee2020co,wang2021explanations,nguyen2018comparing,jung2020limits,dressel2018accuracy,dodge2019explaining,biran2017human}, 
linear regression~\cite{poursabzi2018manipulating,cheng2019explaining,dietvorst2018overcoming,mcgrath2020does},
generalized additive models$^*$ (GAMs)~\cite{abdul2020cogam,springer2018progressive,tan2018investigating,ghai2020explainable,feng2018can,dodge2019explaining,bansal2020does} 
decision trees/random forests~\cite{lee2020co,green2019principles,weerts2019human,zhang2020effect,green2019disparate,lim2009and,tsai2021exploring,chromik2021think,friedler2019assessing,lucic2020does},
support-vector machines (SVMs)~\cite{lee2020co,liu2021understanding,dressel2018accuracy,yin2019understanding,lai2019human,lai2020chicago,ribeiro2016should,yang2020visual}, 
Bayesian decision lists~\cite{lakkaraju2016interpretable},
K-nearest neighbors \cite{kulesza2013too}, shallow (1- to 2-layer) neural networks \cite{nguyen2018comparing,friedler2019assessing}, 
naive Bayes~\cite{smithrenner2020}, matrix factorization~\cite{kunkel2019let}
    \\ 
        \midrule

            Wizard of Oz & \cite{lage2019evaluation,narayanan2018humans,logg2019algorithm,binns2018s,van2021effect,anik2021data,bussone2015role,park2019slow,buccinca2020proxy,buccinca2021trust,lu2021human}
    \\
    \bottomrule
    \end{tabular}
    \caption{Different AI model types used in human-AI decision making, grouped into 3 categories: (1) deep learning models, (2), \shallow models, and (3) Wizard of Oz. We exclude logistic regression and linear regression from GAMS.} 
    \label{tb:models}
    \end{table}

\para{Deep learning models.}
Much recent excitement around AI is driven by the popularity of deep learning models that demonstrate strong performance in a wide variety of tasks and can even outperform humans.
Deep learning models are based on neural networks, which usually consist of more than two layers. 
Deep learning models have been included in many recent studies on human-AI decision making
\cite{alqaraawi2020evaluating,gero2020mental,gonzalez2020human,chandrasekaran2018explanations,hase2020evaluating,guo2019visualizing,kiani2020impact,lee2021human,nourani2021anchoring,cai2019effects,yin2019understanding,cai2019human,peng2019you,bansal2020does,ghazvininejad2017hafez,clark2018creative,lai2020chicago,lee2020co,mcgrath2020does}.
Some papers specified their deep learning models, e.g., \textit{convolution neural networks}~\cite{chandrasekaran2018explanations,alqaraawi2020evaluating}, \textit{recurrent neural networks}~\cite{guo2019visualizing,carton2020feature}, 
\textit{BERT} \cite{lai2020chicago}, and \textit{RoBERTa}~\cite{bansal2020does}, a \textit{hybrid LSTM and CNN} model for VQA task~\cite{ribeiro2018anchors}.
In average training settings, deep learning models typically provide greater predictive power than traditional ``shallow'' models but with the expense of added system complexity.  Deep learning models are commonly considered not directly interpretable and thus raise concerns of user trust.  To tackle this challenge, many ``post-hoc'' explanation techniques~\cite{ribeiro2016should,lundberg2017unified} have been developed to approximate the complex model's logic, which also raise concerns about explanation fidelity~\cite{adebayo2018sanity,sixt2020explanations}. 
We discuss some examples of post-hoc explanation techniques studied in Section~\ref{sec:assistance}.

\para{\Shallow models.} 
Despite the superior performance of deep learning models, many empirical studies employed traditional, \shallow models, which are often easier to train and debug.
These models include \textit{generalized additive models} (e.g., \textit{logistic regression} and \textit{linear regression})~\cite{lee2020co,biran2017human,abdul2020cogam,cheng2019explaining,springer2018progressive,friedler2019assessing,wang2021explanations,nguyen2018comparing,jung2020limits,dressel2018accuracy,tan2018investigating,ghai2020explainable,poursabzi2018manipulating,feng2018can,dodge2019explaining,bansal2020does,dietvorst2018overcoming, mcgrath2020does}, 
\textit{decision trees/random forests}~\cite{lee2020co,green2019principles,weerts2019human,zhang2020effect,green2019disparate,lim2009and,tsai2021exploring,chromik2021think,friedler2019assessing,lucic2020does},
\textit{support vector machines}~\cite{lee2020co,liu2021understanding,dressel2018accuracy,yin2019understanding,lai2020chicago,ribeiro2016should,yang2020visual,lai2019human},
\textit{shallow neural networks}~\cite{nguyen2018comparing,friedler2019assessing},
\textit{Bayesian decision lists}~\cite{lakkaraju2016interpretable},
\textit{K-nearest neighbors} \cite{kulesza2013too}, 
\textit{naive Bayes}~\cite{smithrenner2020}, and \textit{matrix factorization}~\cite{kunkel2019let}.
In prediction tasks with a small number of features, \shallow models are able to achieve competitive accuracy to deep learning models \citep{rudin2019stop}. 
Moreover, some of the simpler \shallow models are deemed to be directly interpretable. For example, coefficients in linear models as feature importance and shallow decision trees are relatively intuitive to comprehend. 
It is worth noting that more papers used shallow models instead of deep learning models to conduct empirical studies on human-AI decision making.

\para{Wizard of Oz.}
Finally, many experiments did not use an actual model, but instead having researchers manually creating and simulating the model output, a common method called ``\textit{Wizard of Oz}'' (WoZ) in HCI research~\cite{kelley1983empirical}. Researchers have used WoZ method with fictional cases of model predictions and explanation styles
~\cite{lage2019evaluation,narayanan2018humans,logg2019algorithm,lu2021human,binns2018s,van2021effect,anik2021data,bussone2015role,park2019slow,buccinca2020proxy,buccinca2021trust}.  WoZ is not only convenient for conducting user studies without investing in technical development, but also gives researchers full control over the interested model behaviors. For instance, utilizing this approach allowed researchers to adjust the algorithm accuracy \cite{park2019slow}, control error types \cite{buccinca2021trust}, and test different explanation styles \cite{bussone2015role,binns2018s,anik2021data,buccinca2020proxy}. However, it can be challenging to design realistic WoZ studies given the complexity of model behaviors. Failing to do so could impair the validity and generalizability of study results. 

\para{Data types}
Besides the models, it is also important to distinguish different types of data used: text, imagery, audio, video, and tabular (or structured data). The surveyed papers used a number of data types, including plain text (e.g., LSAT questions~\cite{bansal2020does}, hotel reviews~\cite{lai2019human,lai2020chicago}, etc.), imagery (potentially cancerous images~\cite{cai2019hello}, meal images~\cite{buccinca2021trust}, etc.), video (stroke patient rehabilitation videos~\cite{lee2021human}, kitchen activity videos~\cite{nourani2021anchoring}, etc.), and tabular (or structured) data (e.g., company financial data~\cite{biran2017human}, personal financial data~\cite{chromik2021think}, etc.). For some tasks, combinations of data types are used. For example, in music recommendations, humans might review structured data (e.g., song title, artist, and genre) and listen to audio when determining whether to listen to a recommended song~\cite{kulesza2012tell}. Video question answering requires assessing both images as well as textual questions about them~\cite{chandrasekaran2018explanations}.

The data type influences what ML models can be applied and how well they perform. For example, shallow models may perform relatively better on structured (or tabular data) than unstructured text compared to deep learning approaches, because of the high dimensionality of text data. 
More importantly, the data type can determine the nature of the decision task and the affordances of AI assistance for the decision. For example, a common form of AI assistance is explanation of the model outputs, such as the models' \textit{attention}. 
For example, prior work found that attention explanations have limited utility for explaining image classifications~\cite{alqaraawi2020evaluating}.
This might be because such explanations---where important areas of the image are painted---can be noisy and confusing to humans compared to attention explanations for text data where important words are highlighted.
Data type can also influence the experience with the decision tasks in many ways: video can take longer to review than images or short text.

\subsection{AI Assistance Elements}
\label{sec:assistance}
\begin{figure}
    \centering
    \includegraphics[width=0.75\textwidth]{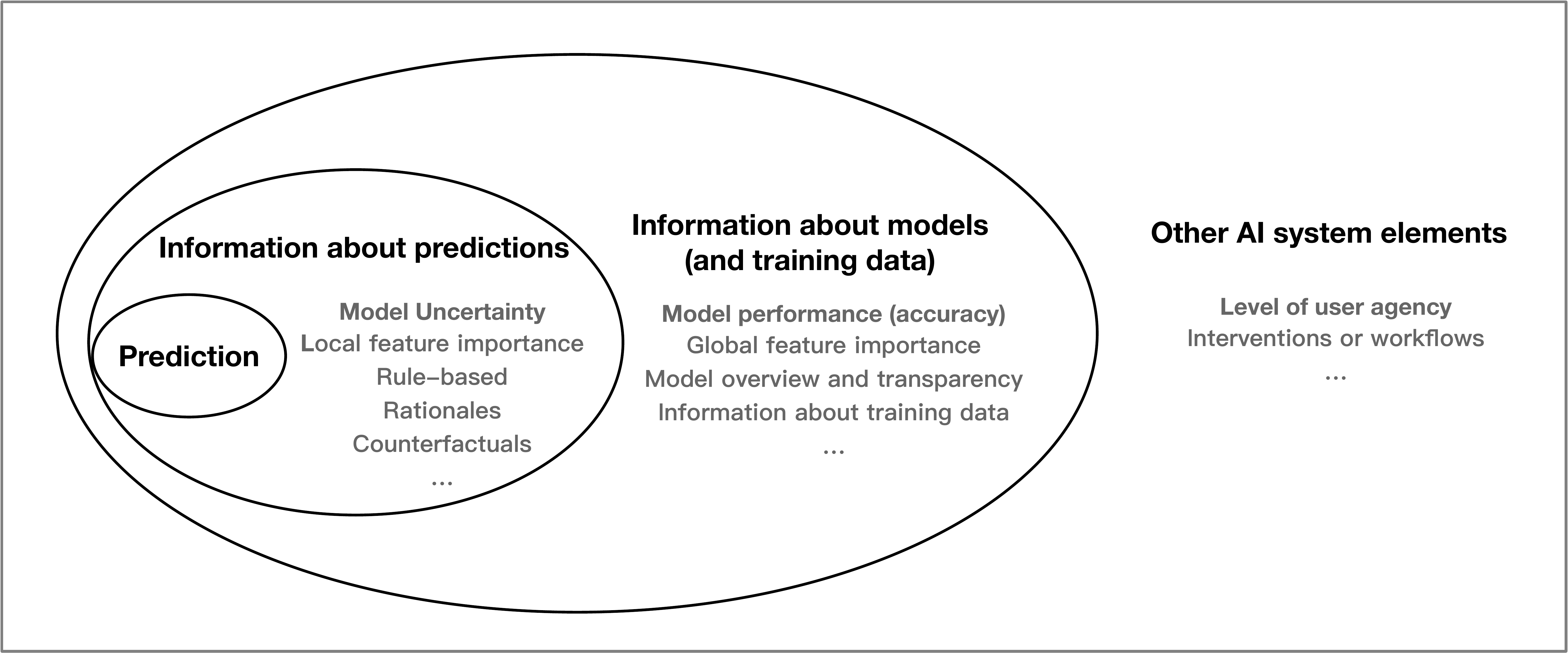}
    \caption{A diagram of AI assistance elements. Systems might simply provide a prediction for which the decision-maker can choose to follow or ignore, or they might provide additional assistance in the form of information about that prediction, information about the model (or training data), or other AI system elements, such as user agency or workflows.}
    \label{fig:assistance_elements}
\end{figure}

A central consideration in studies of human-AI decision making is what kind of assistance is effective in improving decision outcomes.
At the minimum, models can assist decision makers by providing \textit{predictions}, for example,  music or movie recommendations, or generating a health risk scores. 
It is often desirable to provide \textit{information about model predictions} to help users judge whether they should follow them,  especially in cases of disagreement or in high-stakes domains. It is also common to provide \textit{information about the model} to help users gain an overall understanding of how the model works or the data it was trained on, which can influence their perceptions of and interactions with the model in decision making tasks. Figure~\ref{fig:assistance_elements} illustrates our taxonomy of AI assistance elements, which includes predictions, information about predictions, information about models, and other AI system elements that govern the use of the system (e.g., workflows, user control, and varied model quality).

Based on this conceptualization, we categorize the AI assistance elements studied in the survey paper into these four groups, as listed in Table~\ref{tb:assistance}. It is interesting to note that many of the studies reviewed focused on providing information about model predictions or the model in the form of \textit{model explanations}, which are hypothesized to help humans understand model's predictions, detect errors, as well as gain additional information or knowledge about the decision task at hand.

\begin{table}[htbp]

\centering
\resizebox{.5\paperheight}{!}{
\begin{tabular}{p{0.3\textwidth}p{0.65\textwidth}}
\toprule
AI Assistance Elements & Methods \\ 

\midrule
\multicolumn{2}{c}{Model predictions} \\
\midrule
Prediction     
& Binary prediction~\cite{wang2021explanations,liu2021understanding,lu2021human,bansal2019updates,yin2019understanding,zhang2020effect,lai2019human,lai2020chicago,smithrenner2020,binns2018s,dodge2019explaining,bansal2020does,hase2020evaluating,biran2017human,lim2009and,tsai2021exploring,cheng2019explaining,grgic2019human,nourani2021anchoring,yu2019trust}, multi-class~\cite{park2019slow,liu2021understanding,kulesza2013too,bussone2015role,cai2019effects,kunkel2019let,gonzalez2020human,feng2018can,bansal2020does,buccinca2021trust,das2020leveraging,yang2020visual,lee2019procedural,stowers2017insights,guo2019visualizing,ehrlich2011taking,kulesza2012tell,levy2021assessing}, 
continuous prediction or regression~\cite{park2019slow,lee2020co,dietvorst2018overcoming,green2019principles,green2019disparate,logg2019algorithm,dietvorst2015algorithm,de2020case,poursabzi2018manipulating,arshad2015investigating,kiani2020impact,lee2021human,abdul2020cogam,chromik2021think,szymanski2021visual,springer2018progressive}, multiple predictions or multi-step decision~\cite{lage2019evaluation,narayanan2018humans,lee2020co,lee2021human,buccinca2021trust,lee2019procedural,levy2021assessing,kiani2020impact}, with alternates~\cite{feng2018can,bansal2020does,stowers2017insights,guo2019visualizing,ehrlich2011taking,arshad2015investigating}\\
No prediction shown & \cite{friedler2019assessing,lakkaraju2016interpretable,lucic2020does,gero2020mental,bansal2019updates,bansal2019beyond,weerts2019human,nguyen2018comparing,harrison2020empirical,rader2018explanations,cai2019human,chandrasekaran2018explanations,kocielnik2019will,van2021effect,anik2021data,alqaraawi2020evaluating}
\\
\midrule
\multicolumn{2}{c}{Information about model predictions} \\
\midrule
Model uncertainty & Classification confidence (or probability) \cite{weerts2019human,jung2020limits,zhang2020effect,feng2018can,bansal2020does,buccinca2021trust,guo2019visualizing,bussone2015role,arshad2015investigating,kiani2020impact,lee2020co,lee2021human}, flag low confidence~\cite{levy2021assessing}, uncertainty distribution (regression)~\cite{mcgrath2020does} 
\\
Local feature importance        
& 
Coefficients \cite{biran2017human,springer2018progressive,cheng2019explaining,liu2021understanding,wang2021explanations,green2019principles,nguyen2018comparing,smithrenner2020,lai2020chicago,feng2018can,dodge2019explaining,ghai2020explainable,lai2019human,poursabzi2018manipulating,lee2020co,lee2021human}, 
attention \cite{lai2020chicago,chandrasekaran2018explanations,carton2020feature}, 
gradient-based~\cite{nguyen2018comparing,kiani2020impact,chandrasekaran2018explanations},
propagation-based (LRP~\cite{alqaraawi2020evaluating}),
perturbation-based 
(LIME ~\cite{nguyen2018comparing,hase2020evaluating,alqaraawi2020evaluating,bansal2020does}, 
SHAP~\cite{weerts2019human,zhang2020effect,chromik2021think}),
Wizard of Oz~\cite{binns2018s,bussone2015role,buccinca2021trust,buccinca2020proxy},
video features~\cite{nourani2021anchoring}\\

Rule-based explanations 
& 
Decision sets~\cite{lage2019evaluation,narayanan2018humans,lakkaraju2016interpretable}, tree-based explanation~\cite{kulesza2013too,lim2009and},
anchors~\cite{ribeiro2018anchors,hase2020evaluating}
\\
Example-based methods 
& 
Nearest neighbor or similar training instances~\cite{cai2019human,binns2018s,cai2019effects,kulesza2013too,kunkel2019let,buccinca2020proxy,wang2021explanations,dodge2019explaining,lai2019human,tsai2021exploring,hase2020evaluating}
\\

Counterfactual explanations                  
& 
Contrastive or sensitive features~\cite{dodge2019explaining,lucic2020does}, counterfactual examples~\cite{friedler2019assessing,binns2018s,wang2021explanations,lim2009and}
\\
Natural language explanations 
& 
Model-generated rationales~\cite{das2020leveraging,tsai2021exploring,biran2017human}, expert-generated rationales~\cite{bansal2020does}
\\
Partial decision boundary 
& 
Traversing the latent space around a data input~\cite{hase2020evaluating}
\\
\midrule
\multicolumn{2}{c}{Information about models (and training data)} \\
\midrule
Model performance & Accuracy~\cite{yang2020visual,yin2019understanding,lai2019human,lai2020chicago,harrison2020empirical}, false positive rates~\cite{harrison2020empirical} \\
Global feature importance       & 
Coefficients~\cite{dodge2019explaining}, 
permutation-based~\cite{wang2021explanations}, 
shape function of GAMs~\cite{abdul2020cogam}, 
Wizard of Oz~\cite{binns2018s} \\
Presentation of simple models
& 
Decision sets~\cite{lakkaraju2016interpretable},
decision trees~\cite{friedler2019assessing}, 
linear regression~\cite{poursabzi2018manipulating}, 
logistic regression~\cite{friedler2019assessing}, 
one-layer MLP~\cite{friedler2019assessing}
\\
Global example-based explanations 
& 
Model tutorial~\cite{lai2020chicago}, prototypes~\cite{cai2019effects,feng2018can,kocielnik2019will}
\\
Model documentation 
& 
Overview of the model or algorithm~\cite{lee2019procedural,rader2018explanations,kulesza2013too,kulesza2012tell,kocielnik2019will}, 
model prediction distribution~\cite{van2021effect}  \\
Information about training data 
& 
Input features or information the model considers~\cite{kulesza2013too,poursabzi2018manipulating,harrison2020empirical,zhang2020effect}, aggregate statistics (e.g., demographic)~\cite{binns2018s,dodge2019explaining}, full training ``data explanation''~\cite{anik2021data}
\\\midrule

\multicolumn{2}{c}{Other AI system elements affecting user agency or experience} \\
\midrule
Interventions or workflows affecting cognitive process &
User makes prediction before model~\cite{lu2021human, zhang2020effect, grgic2019human,wang2021explanations,yin2019understanding,poursabzi2018manipulating, buccinca2021trust}, 
vary system response times~\cite{park2019slow, buccinca2021trust},
outcome feedback to user~\cite{chandrasekaran2018explanations, grgic2019human, yu2019trust, yang2020visual, bansal2019updates,bansal2020does}, 
training phase~\cite{lai2020chicago,chandrasekaran2018explanations, zhang2020effect,poursabzi2018manipulating,kulesza2012tell}, 
source of recommendation or local explanation (human or AI)~\cite{kunkel2019let,logg2019algorithm,dietvorst2015algorithm},
varied model quality~\cite{levy2021assessing, smithrenner2020, ehrlich2011taking, nourani2021anchoring, yu2019trust, kocielnik2019will,park2019slow,yin2019understanding} \\
Levels of user agency  & Allowing user feedback or personalization for model~\cite{feng2018can, smithrenner2020,kulesza2012tell}, outcome control (after decision~\cite{lee2019procedural, zhang2020effect,dietvorst2018overcoming}, before decision~\cite{kocielnik2019will}), 
interactive explanations~\cite{cai2019human,liu2021understanding,cheng2019explaining}, 
user direction on input data~\cite{kiani2020impact, levy2021assessing,cai2019human}, level of machine agency~\cite{levy2021assessing,buccinca2021trust}
\\
\bottomrule
\end{tabular}
}
\caption{AI assistance elements explored in prior work on human-AI decision making, grouped into four categories: (1) model predictions, (2) information about model predictions, (3) information about models (and training data), and (4) other AI system elements affecting user agency or experience.}
\label{tb:assistance}
\end{table}

\subsubsection{Model Predictions}
Providing a model's prediction is the most natural form of assistance for decision making. 
For example, in medical decision support tasks, models can predict medical diagnoses based on patient information; here, a doctor can decide whether to follow or ignore the prediction from the model in favor of a different diagnosis. 

We can differentiate AI assistants based on the types of predictions they provide, including \textit{binary predictions} (e.g., positive or negative sentiment or binary recidivism prediction)~\cite{wang2021explanations,liu2021understanding,lu2021human,bansal2019updates,yin2019understanding,zhang2020effect,lai2019human,lai2020chicago,smithrenner2020,binns2018s,dodge2019explaining,bansal2020does,hase2020evaluating,biran2017human,lim2009and,tsai2021exploring,cheng2019explaining,grgic2019human,nourani2021anchoring,yu2019trust}, \textit{multi-class}, or categorical predictions (e.g., object type in an image or song recommendation)~\cite{park2019slow,liu2021understanding,kulesza2013too,bussone2015role,cai2019effects,kunkel2019let,gonzalez2020human,feng2018can,bansal2020does,buccinca2021trust,das2020leveraging,yang2020visual,lee2019procedural,stowers2017insights,guo2019visualizing,ehrlich2011taking,kulesza2012tell,levy2021assessing}, and \textit{continuous predictions or regressions} (e.g., house prices)~\cite{park2019slow,lee2020co,dietvorst2018overcoming,green2019principles,green2019disparate,logg2019algorithm,dietvorst2015algorithm,de2020case,poursabzi2018manipulating,arshad2015investigating,kiani2020impact,lee2021human,abdul2020cogam,chromik2021think,szymanski2021visual,springer2018progressive}. 
Some systems provide multiple types of predictions or perform \textit{multi-step decision making}~\cite{lage2019evaluation,narayanan2018humans,lee2020co,lee2021human,buccinca2021trust,lee2019procedural,levy2021assessing,kiani2020impact}.
For example, the calorie-cutting recipe system of \citet{buccinca2021trust} performs a multi-step decision, first choosing the highest calorie food and then predicting replacement alternatives for it. 
And, while the majority of systems show only the top prediction, some provide prediction alternates as additional options for users to review~\cite{feng2018can,bansal2020does,stowers2017insights,guo2019visualizing,ehrlich2011taking,arshad2015investigating}.
 
It is worth noting that some papers did not study AI assistance with predictions~\cite{friedler2019assessing,lakkaraju2016interpretable,lucic2020does,gero2020mental,bansal2019updates,bansal2019beyond,weerts2019human,nguyen2018comparing,harrison2020empirical,rader2018explanations,cai2019human,chandrasekaran2018explanations,kocielnik2019will,van2021effect,anik2021data,alqaraawi2020evaluating}. 
One reason is that they chose to focus on evaluating people's perceptions of the model based on other information, without implementing the complete decision making process. 
For example, \citet{chandrasekaran2018explanations,buccinca2020proxy} ask humans to simulate the prediction or predict its correctness based on the explanations. \citet{anik2021data} asks participants to judge the model's fairness and trust based on data-centric explanations.
Another reason for not revealing model predictions is to retain as much human agency as possible in the decision making process \citep{lai2019human}.

\subsubsection{Information about Model Predictions}
Information about predictions can include the model's uncertainty for the prediction and other prediction-specific explanation (or \textit{local explanations}), which help users understand why the prediction was made. 
Such local explanations include local feature importance, rule-based explanations, example-based explanations, counterfactual explanations, and natural language explanations.
For our review, we primarily focus on the \textit{forms} of explanation and how they are presented to humans instead of the underlying algorithm/computation. 

\para{Model uncertainty.}
Showing the model's \textit{uncertainty} for a prediction can make humans aware of when the model is more or less sure, so long as uncertainty estimates are reliable~\cite{ghosh2021uncertainty}. 
In theory, a low uncertainty should alert users to not over-rely on the prediction and resort to their own judgment or other resources.

The most common form of uncertainty information is a confidence score or prediction probability for a classification model. 
Here, uncertainty is usually calculated as the probability, a numeric value between 0 to 1, associated with the predicted label (opposed to alternative labels) given by the underlying ML model. 
Many prior systems expose \textit{classification confidence} scores to human decision makers~\cite{weerts2019human,jung2020limits,zhang2020effect,feng2018can,bansal2020does,buccinca2021trust,guo2019visualizing,bussone2015role,arshad2015investigating,kiani2020impact,lee2020co,lee2021human}.
Among these studies, \citet{bussone2015role,buccinca2021trust} generate confidence scores using ``Wizard of Oz'', while the others use the classification model to generate probabilities and typically present these scores alongside the prediction.
Other classification systems expose uncertainty with labeled categories. For example, \citet{levy2021assessing} label some predictions as low confidence in one version of their clinical notes annotation system.

In contrast to classification, \textit{uncertainty information of regression} models is currently under-explored in human-AI decision making, with one recent exception~\cite{mcgrath2020does} (though the effect of uncertainty information on decision making has long been studied outside the context of AI assistance, e.g.~\cite{kay2016ish}).
Uncertainty for regression models can take the form of uncertainty distribution---how the possible values are distributed (often centered around the given prediction)---or prediction interval---the range of possible values.
  
Currently there is also a lack of discussion on the reliability, or sometimes referred to as calibration~\cite{ghosh2021uncertainty}, of uncertainty estimates, and whether decision makers can make sense of uncertainty estimates properly.  
For example, in some deep learning models, prediction probabilities are prone to overconfidence~\cite{guo2017calibration}.
Recent work has experimented with deep probabilistic models to give more reliable uncertainty estimations, including Bayesian neural networks~\cite{Zhang2019AdvancesIV,tran2019bayesian}, deep neural networks that integrate dropout~\cite{gal2016dropout}, or ensemble methods to approximate Bayesian inference~\cite{lakshminarayanan2017simple}. 
How reliability of uncertainty information affects decision making, or how to communicate the reliability or a lack thereof, remain open questions in the context of human-AI decision making.

\para{Local feature importance.} 
\label{sec:assistance:local-feature-importance}
Local explanation techniques provide information about \textit{how} and \textit{why} a given prediction is made to assist humans' judgment of the prediction and inform their final decision.
A common local explanation type is local feature importance, which, 
given a single instance, quantifies the contribution (or importance) of each of its features to the model's prediction of it. For example, when predicting property values, certain features are more important to the prediction (e.g., lot size and number of rooms) while others might be less important (e.g., distance to a school). 
\citet{tsai2021exploring} use feature importance to explain the diagnosis for a COVID-19 chatbot (e.g., ``I [provided] this recommendation [because]: 1) you are feeling ill, 2) you have COVID-19 related contact history...). 
We enumerate the common ways to derive local feature importance grouped by built-in, post-hoc, and Wizard of Oz techniques:

\begin{itemize}
    \item \textbf{Built-in methods}, including coefficients and attention, are common ways to generate local feature importance directly from the models. Models that can generate feature importance explanations from their \textit{coefficients} are often considered as ``directly interpretable''.
    For instance, the coefficient for GAMs is a direct measure of feature importance \cite{schielzeth2010simple}, when each feature is scaled to have the standard deviation.
    Many papers~\cite{biran2017human,springer2018progressive,cheng2019explaining,liu2021understanding,wang2021explanations,green2019principles,nguyen2018comparing,smithrenner2020,lai2020chicago,feng2018can,dodge2019explaining,ghai2020explainable,lai2019human,poursabzi2018manipulating,lee2020co,lee2021human}
    adopted coefficients as the feature importance scores.
    Second, \textit{attention} mechanism is a common component in deep learning models, especially in the context of NLP. Attention as explanations shows a subset of input features that the deep learning model should pay more attention to for a particular prediction. 
    \citet{lai2020chicago,chandrasekaran2018explanations,carton2020feature} adopt attention mechanism to model the local feature importance.\footnote{There are recent debates on whether attention is explanations. \citet{serrano2019attention,jain2019attention} raise concern that attention might not exclusively lead to predicted label. \citet{wiegreffe2019attention} argue that attention could be explanations depending on the definition of explanations.}
    
    \item \textbf{Post-hoc methods} learn to generate explanations separately for a trained model, often a non-interpretable model such as deep neural networks. These methods can be  grouped into:
    gradient-based~\cite{nguyen2018comparing,kiani2020impact,chandrasekaran2018explanations},
    propagation-based (LRP~\cite{alqaraawi2020evaluating}),
    and perturbation-based 
    (e.g., LIME~\cite{nguyen2018comparing,hase2020evaluating,alqaraawi2020evaluating,bansal2020does}, 
    SHAP~\cite{weerts2019human,zhang2020effect,chromik2021think}).
    First, \textit{gradient-based methods}  compute the gradient of the prediction with respect to the input features. Examples using gradient-based methods include classification activation map (CAM)~\cite{kiani2020impact} and Grad-CAM~\cite{chandrasekaran2018explanations}. 
    Second, \textit{propagation-based methods}, especially Layer-wise Relevance Propagation (LRP), uses a forward pass and then a backward pass to calculate the relevance among input features. \citet{alqaraawi2020evaluating} adopted LRP in their experiments.
    Third, \textit{perturbation-based methods}, such as SHAP and LIME, manipulates parts of the inputs to generate explanations. 
    LIME \cite{ribeiro2016should} uses a sparse linear model to approximate the behavior of a machine learning model locally. The coefficients of this sparse linear model can then serve as explanations.
    \citet{nguyen2018comparing,hase2020evaluating,alqaraawi2020evaluating,bansal2020does} use explanations from LIME as AI assistance.
    SHAP (SHapley Additive exPlanations) provides the marginal contribution of each feature for a particular prediction, averaged over all possible permutations, which is first proposed by \citet{lundberg2017unified}.  Shapley values assign each feature an importance value for a particular prediction. In the context of human-AI decision making, \citet{weerts2019human,zhang2020effect} use SHAP for local feature importance.
    In addition, deep learning models to generate video-specific feature captions~\cite{nourani2021anchoring} are also being used. 

    \item {\bf Wizard of Oz} allows manual creation of local feature importance. 
    Researchers often choose to adopt \textit{Wizard-of-Oz} to simulate both the AI model predictions and their explanations. Wizard-of-Oz is often chosen to save technical investment and have more control over studying interested AI model behaviors and explanation styles, as mentioned in \citet{binns2018s,bussone2015role,buccinca2021trust,buccinca2020proxy}. However, they can deviate from the actual output of existing explanation methods, and researchers should carefully justify the design choices and consider the limitations.
\end{itemize}

\para{Rule-based explanations.}
Rule-based explanations are constructed with a combination of rules, where a rule can be a simple `if-then' statement. 
Both built-in approaches (e.g., decision sets and decision trees) and post-hoc approaches (e.g., anchors) have been explored for generating rule-based explanations in the context of human-AI decision making.

\begin{itemize}
    \item \textbf{Decision sets}: \citet{lakkaraju2016interpretable} generate interpretable \textit{decision sets}, which are sets of if-then rules to explain model decisions. In their study, participants are asked to describe the characteristics of certain classes (e.g., depression) based on the learned decision set for that class. 
    \citet{lage2019evaluation,narayanan2018humans} also used decision sets as local explanations in their studies.
    \item \textbf{Tree-based explanations}: 
    For \textit{decision tree-based} models, local explanations can be generated directly by the decision-tree path, as rules that the model followed, to reach the given decision. Tree-based explanations were used by \citet{kulesza2013too} in the context of a music recommendation system, which employed a content-based decision tree approach for selecting songs. \citet{lim2009and} used the underlying decision tree model to generate multiple types of explanations (\textit{why}, \textit{why not}, \textit{how to}, and \textit{what if}), such as by the decision-tree paths to reach an alternative decision.
   
    \item \textbf{Anchors}, proposed by \citet{ribeiro2018anchors}, learns if-then rules representing ``sufficient'' conditions (important features) that guarantee the given input to have the prediction, such that changes to the rest of the features will not change the prediction.  \citet{ribeiro2018anchors,hase2020evaluating} explored anchors as an explanation methods in their experiments.
\end{itemize}

\para{Example-based methods.}
Example-based explanation methods explains a prediction by examples (with known outcomes) to support case-based reasoning. 
A common formulation is to find instances from training dataset that are similar to the given input.
The explanations should include their labels in the ground truth to help users make sense of reasons behind the current prediction.  
For example, we might explain the predicted price for a given home by showing similar homes with their actual prices.
A common and simple way to generate \textit{similar instances} is to find the \textit{nearest neighbors} in the embedding (latent representation) space. 
This method is used by many papers to explain predictions for human-AI decision making~\cite{cai2019human,binns2018s,cai2019effects,kulesza2013too,kunkel2019let,buccinca2020proxy,wang2021explanations,dodge2019explaining,lai2019human,tsai2021exploring,hase2020evaluating}.

\para{Counterfactual Explanations.}
Counterfactual explanations help people understand how the current input should change to get an alternative prediction, answering ``why not'' (a different prediction) or ``how to be a different prediction'' instead of a ``why'' question. Counterfactual explanations can also be provided based on either features or examples. Feature-based ones are often called \textit{contrastive feature} or sensitive feature methods---highlighting features that if changed, often implying minimum change, will alter the prediction to the alternative class. For example, a counterfactual explanation for loan prediction task could be ``you would have received the loan if your income was higher by \$10,000.'' Example-based are often called \textit{counterfactual examples}, by providing examples in the training data or hypothetical examples with the alternative class label but minimum differences from the current input. 

A small but growing number of prior works studying human-AI decision making have utilized counterfactual explanations, such as contrastive and sensitive features~\cite{dodge2019explaining,lucic2020does}.
For example, \citet{dodge2019explaining} present sensitivity-based explanations to help people discover unfair decisions (by changing just the race feature, the model would have a different prediction).
And, \citet{lucic2020does} generate contrastive explanations for model errors %
by identifying feature ranges that lead to reasonable predictions 
Similarly, other work uses counterfactual examples include ~\cite{friedler2019assessing,binns2018s,wang2021explanations,lim2009and}.
For example, \citet{wang2021explanations} show instances with minimal changes that result in the desired output. \citet{friedler2019assessing} asked users to answer `what if' question given a perturbed input. 

\para{Natural language explanations.}
Natural language explanations, or sometimes referred to as \textit{rationale-based explanations}, are a form of ``why'' explanations that provide the reasoning or \textit{rationale} behind a particular decision. 
For example, \citet{tsai2021exploring} study rationale-based explanations for their COVID-19 chatbot, such as why the chatbot asks particular diagnostic questions to the user. 
These explanation types are sometimes referred to as ``justifications''~\cite{biran2017human}. 
Natural language explanations can be differentiated by how they are generated, either \textit{model/algorithm generated}~\cite{das2020leveraging,tsai2021exploring,biran2017human}---where these explanations are produced by the system---or \textit{human experts generated}~\cite{bansal2020does}, meaning domain experts (or algorithm developers) provided rationales behind types of predictions to be shown to users.

\para{Partial decision boundary.}
\citet{hase2020evaluating} showed the model's partial decision boundary by \textit{traversing the latent space around a specific input}, in order to show how the model behaves as the input changes. The methods were initially proposed in the computer vision domain~\citep{joshi2018xgems,samangouei2018explaingan}, whereas ~\citet{hase2020evaluating} developed and adapted the method for text and tabular data.

\subsubsection{Information about models (including training data)}
Users of AI systems are often interested in better understanding the underlying model to form an appropriate mental model that can help them interact more effectively. ``Global'' information about the model can include the model's overall performance, global explanations (e.g., how the model weighs different features, visualizing the whole model processes for simple models), input and output spaces, information about the training data, provenance, and more. Recently, there are growing interests in providing documentation or `About Me'' page to present such global information (e.g., Model cards~\cite{mitchell2019model}, FactSheets~\cite{arnold2019factsheets}). In this section, we discuss what types of global information about models have been studied in surveyed papers of human-AI decision making.

\para{Model performance.}
\label{sec:assistance-global-performance}
Model performance describes how well a model works in general.
In studies of human-AI decision making, model performance has been mainly presented  in the form of \textit{accuracy} (i.e., percentage of correctly predicted instances)~\cite{yang2020visual, yin2019understanding,lai2019human,lai2020chicago,harrison2020empirical}.
These works typically explore how observing model accuracy affects people's perception of and decision making with the model.
For example, \citet{lai2019human} investigate whether human subjects' awareness of the ML model's accuracy improves their performance in decision making tasks.
And, \citet{yin2019understanding} studies the effect of accuracy on human's trust in ML models.

Model performance has also been described by \textit{false positive rates}, or how frequently the system mislabels an input as a particular class.
For example, \citet{harrison2020empirical} showed the presenting false positive rates in addition to accuracy in their experiments helped people gauge fairness of the model for recidivism prediction tasks.

It is useful to note that accuracy information is usually estimated on a held-out dataset, and the model's actual performance in deployment can shift, especially when the actual decision inputs or their distribution differ from that of the training data. This gap between communicated accuracy and experienced accuracy has been studied in~\citet{yin2019understanding}. Future work should also explore the effects of other types of performance metrics, such as precision and recall.

\para{Global feature importance.}
Different from local feature importance that quantifies each feature's importance to \textit{a specific prediction}, global feature importance quantifies each feature's overall importance to \textit{the model's decisions for a given task}.
Here we enumerate methods used in surveyed papers for computing global feature importance, grouped into built-in methods, post-hoc methods, and Wizard of Oz as follows:
\begin{itemize}
    \item \textbf{Built-in methods} derive the global feature importance score directly from the trained model. 
    Two such examples are coefficients and shape function of GAMs.
    \textit{Coefficients} derived from a logistic regression model encode the relative importance of each feature.
    For example, \citet{dodge2019explaining} use coefficients to present the global influence of features by showing their positive or negative effect on the decision boundary.
    The {\it shape function of GAMs} 
    is used to inspect the global feature importance for GAMs. 
    Mathematically, GAM is modeled as $y=\beta_0 + f_1(x_1) + f_2(x_2)+....+f_p(x_p)$, 
    where $y$ denotes the prediction and $x_i$ denotes the input feature. 
    The shape function $f_i$ describes the global importance of feature $x_i$.
    \citet{abdul2020cogam} visualizes the shape function via charts that serve as global explanations of their system.
    \item \textbf{Post-hoc methods} generate global feature importance in a post-hoc manner for a trained model, often for a complex, non-interpretable model. 
    One such method is \textit{permutation importance}, also known as variable importance (VI), which measures the increase in model prediction error when an input feature is permuted. 
    Early work used permutation importance to compute global feature importance for random forests~\citet{breiman2001random} followed by a rich line of research on the topic~\cite{strobl2008conditional,altmann2010permutation,zhu2015reinforcement,gregorutti2015grouped,datta2016algorithmic,gregorutti2017correlation}. 
    More recently, \citet{fisher2019all} proposed a model-agnostic version of permutation importance. \citet{wang2021explanations} adopt this method \citet{fisher2019all} to compute global feature importance in their paper, exploring whether such explanations are helpful during decision making tasks.
    \item \textbf{Wizard of Oz} manually constructs the global feature importance. For example, \citet{binns2018s} created scenarios of recidivism prediction with hypothetical global feature importance to explore people's fairness perception. 
\end{itemize}

\para{Presentation of simple models.}
For simple models, it is possible to present the whole or part of the model internals to humans to give them a detailed view of how the model makes decisions.
These simple---often referred to as \textit{inherently transparent} or \textit{intrinsically interpretable}---models can be presented to humans in the form of decision tree, rule sets, graphs, or other visualizations. 
For this reason, such models are often preferred over more complicated architectures (e.g., neural nets) when interpretability is desired.

In the context of human-AI decision making, researchers have explored presentations of simple models, including \textit{decision sets}~\cite{lakkaraju2016interpretable} and \textit{trees}~\cite{friedler2019assessing}, \textit{linear}~\cite{poursabzi2018manipulating} and \textit{logistic regressions}~\cite{friedler2019assessing}, and \textit{one-layer multilayer perceptron} (MLP)s~\cite{friedler2019assessing}.
For example, \citet{lakkaraju2016interpretable} construct a small number of compact decision sets that are capable of explaining the behavior of blackbox models in certain parts of feature space. 
\citet{friedler2019assessing} compare three models, representing a decision tree as a node-link diagram and both a logistic regression and a one-layer MLP as math worksheets that are intended to ``walk the users through the calculations without any previous training in using the model.''

\para{Global example-based explanations.}
Example based explanations are instances from the training set that explain the prediction or provide insights to the data which would help humans make task decisions.
\citet{lai2020chicago} select examples with features that provide great coverage from training set as \textit{tutorial} to the task using the SP-LIME algorithm \cite{ribeiro2016should}.
They also proposed the Spaced Repetition algorithm that creates a set of examples which exposes humans to important features repeatedly.
Another common approach is to pinpoint one or a set of training samples that are representative of prototypical instance (with the outcome of a given prediction class)~\cite{nguyen2016synthesizing}. The representative data instance is called a \textit{prototype}. For linear models, it is natural to find the important training example based on the distance in the representation space~\cite{papernot2018deep}. 
For nonlinear models, influence function \cite{koh2017understanding} and representer value \cite{yeh2018representer} are proposed.
\citet{kocielnik2019will} use Wizard of Oz to generate a table of representative instances for each prediction class respectively to help user understand how the AI component operates.
Note that prototypical examples can also be used to explain a prediction locally, by presenting the prototype in its proximity, such as the explanations used in 
 \citet{cai2019effects,feng2018can}.

\para{Model documentation.}
The requirements for model documentation or ``About Me'' page, which provides not only an \textit{overview of the model} characteristics but also how it is developed and intended to be used, are discussed in recent literature as critical to AI transparency and governance ~\cite{mitchell2019model,arnold2019factsheets}. However, only a small number of surveyed studies explored using relevant features in human-AI decision making~\cite{lee2019procedural,rader2018explanations,kulesza2013too,kulesza2012tell,kocielnik2019will}.
For example, the meeting scheduling assistant of \citet{kocielnik2019will} includes a description of how the scheduling assistant works, specifically, ``The Scheduling Assistant examines each sentence separately and looks for meeting related phrases to make a decision.'' 

Some work on model documentation argued for the importance of providing an overview of model's input and output spaces, such as the \textit{output distribution}. \citet{van2021effect} display the race and sex group filters based on the participant’s demographic information (e.g., a male participant first sees the data of male loan applicants).

\para{Information about training data.} 
Finally, human-AI decision making systems can help people better understand the models by providing information about the data on which they were trained, such as the \textit{inputs features} used or data distribution ~\cite{kulesza2013too,poursabzi2018manipulating,harrison2020empirical,zhang2020effect,dodge2019explaining}.
For example, in the income prediction system of \citet{zhang2020effect}, humans are made aware of whether or not the model considers ``marital status'' as a feature.
Some studies present demographic-based or \textit{aggregated statistics} of the training data~\cite{dodge2019explaining,binns2018s}. Finally, \citet{anik2021data}, in a more detailed way, presents a \textit{`full' training data explanation}, including how the data was collected. They also describes demographics, recommended usage, potential issues, and so on.

\subsubsection{Other AI System Elements Affecting User Agency or Experience}

Besides providing information about the AI to assist decision making, prior research also studied additional system elements that can affect user experience, mainly around interventions that affect users' cognitive processes of decision making, or users' agency over the system.

\para{Interventions or workflows affecting cognitive processes.}
Besides providing information about the model and predictions, how people process such information to form perceptions of AI and make decisions can be impacted by interventions that change their cognitive processes

One area of interventions is concerned with how to design the workflow, such as \textit{when} users make their own decisions versus seeing the model's predictions.   
The typical paradigm of human-AI decision making is to have models providing predictions, then users can choose to follow or ignore. 
Some studies explored having users \textit{making their own predictions before being shown the model output}~\cite{lu2021human, zhang2020effect, grgic2019human,wang2021explanations,yin2019understanding,poursabzi2018manipulating, buccinca2021trust}. 
Such designs force people to engage more deliberately with their own decision making rather than relying on the model predictions.

Prior work also explored the impact of different workflow design on users' mental models of how AI assistants make decisions. 
Some systems include a \textit{training phase} prior to the task, during which users review model outputs and explanations of how the system works~\cite{chandrasekaran2018explanations, zhang2020effect,poursabzi2018manipulating,kulesza2012tell,lai2020chicago}. 
In some real-world scenarios, decision makers can see the actual outcomes of decisions. 
Studies have also explored how receiving \textit{outcome feedback} on either their or the models' decision correctness~\cite{chandrasekaran2018explanations, grgic2019human, yu2019trust, yang2020visual, bansal2019updates,bansal2020does} impact people's perception of the models and performance of the tasks.

Another type of intervention studied is \textit{system response time}, or how long the system takes to provide a decision~\cite{park2019slow,buccinca2021trust}. 
For example, \citet{buccinca2021trust} compare the effect of cognitive forcing functions, where participants get suggestions immediately as opposed to having to wait 30 seconds for the machine's prediction, on over-reliance and subjective perception.

The models' actual performance (as opposed to communicated performance metrics described in Section~\ref{sec:assistance-global-performance}) can govern the usefulness of the decision support. 
Prior work also explored how \textit{varying model performance} or prediction quality impacts human-AI decision making ~\cite{levy2021assessing, smithrenner2020, ehrlich2011taking, nourani2021anchoring, yu2019trust, kocielnik2019will,park2019slow,yin2019understanding}.
For example, \citet{smithrenner2020} explored whether describing model errors, which enables users to gauge model performance, without the opportunity to make fixes yielded user frustration for both low and high quality models.  Similarly, \citet{kocielnik2019will} studied the difference between high precision and high recall models (without explicitly showing this information to users) on user perceptions.

Lastly, some studies looked at how the \textit{source of assistance}, whether from an AI versus from a human, affect decision making~\cite{kunkel2019let,logg2019algorithm,dietvorst2015algorithm}. 
For example, \citet{kunkel2019let} explore how machine generated versus human generated explanation impact the acceptance and trust of a system for movie recommendations.

\para{Levels of user agency.}
Typical decision-support AI systems work in a closed loop without the possibilities for guidance from end users. This kind of set-up limits the agency users can have for controlling or improving the decision assistance from the AI.
Some studies have explored improving the user agency, such as allowing and incorporating \textit{user feedback on predictions or personalization of the model}~\cite{feng2018can, smithrenner2020,kulesza2012tell}.
For example, in the music recommendation system of \citet{kulesza2012tell}, participants can provide feedback about the recommended songs and guidelines to the model to improve future recommendations.
Another studied form of user agency is the \textit{ability to guide the prediction (or outcome)} either before~\cite{kocielnik2019will} or after the model's decisions~\cite{lee2019procedural, zhang2020effect, dietvorst2018overcoming}.
For example, \citet{kocielnik2019will} explore user experience with a system for detecting meeting requests from emails. They compare whether providing users a slider to control whether the system tends towards false positive (high recall) or false negatives (high precision) improves experience. This control occurs before the system makes predictions, but can be updated as needed. 
 \citet{lee2019procedural} study whether allowing participants to override the outputs of a system on how to split food between grad students promotes fairness perception.

More recently, \textit{interactive explanations} 
have been investigated, which allow humans to have better control on what kind of explanations they can get from the model~\cite{cai2019human,liu2021understanding,cheng2019explaining}.
For example, \citet{cai2019human} propose and evaluate an interactive refinement tool to extract similar images for pathologists in medical domain. In their tool, users are not provided with explicit explanations, but instead can interact with the system and test it under different hypotheses.  
\citet{cheng2019explaining} compare different explanation interfaces for understanding an algorithm's university admission decisions. They find that an interactive explanation approach is more effective at improving comprehension compared to static explanations, but at the expense of taking more time.

Another form of user control in human-AI decision making is to allow users to choose input data to get model predictions.  Prior work explores cases where \textit{users choose the input data}a (or the underlying features) for the model to consider~\cite{kiani2020impact, levy2021assessing,cai2019human}.
For example, in the similar image search system of \citet{cai2019human}, participants denote the important points of the initial image for the system to attend to when looking for similar images. 

Finally, researchers compare different \textit{levels of machine agency}. For example, \citet{levy2021assessing} experimented with two distinct clinical note annotation systems: one only suggests annotation labels after users choose text spans to be labeled, and another performs both span and label suggestions.
Similarly, \citet{buccinca2021trust} examined receiving \textit{predictions only on demand}.

\subsection{Summary and Takeaways}
We summarize current trends and gaps in how AI models and assistance elements are used and studied, then make recommendations for future work.

\paragraph{Current trends.}

\begin{enumerate}
    \item \textbf{Limited uses of deep learning models.} Despite the popularity of deep learning models, a large proportion of empirical studies still adopted traditional shallow models, even Wizard of Oz, possibly due to their ease to develop or access. Further, shallow models are typically less complicated to explain, making them an easier choice for studying human-AI decision making assisted by information about the prediction or model.  
    \item \textbf{Assistance beyond predictions.} Besides the prediction itself, empirical studies have explored the effect of a wide range of elements providing information about the prediction and the model on improving decision performance. By summarizing these elements studied, we hope to also inform the design space of AI decision support systems.
    \item \textbf{A focus on AI explanations.} A large portion of prior work focused on studying the effect of explanations, both local explanations for a prediction and global explanations for the model.  This is necessary information for people to better understand the AI to interact, but also partly due to a recent surge in the field of explainable AI (XAI), which produces increasing technical availability to generate explanations.
    
    \item \textbf{Beyond the model.} Beyond model-centric assistance elements, a small portion of work explored system elements that affect user agency and action spaces, including workflow and user control.
\end{enumerate}

\paragraph{Gaps in current practices.}

\begin{enumerate}
    \item \textbf{A fragmented understanding and limited design space of AI assistance elements.} Current human-subjects studies often focus on one or a small set of AI assistance elements. We have very limited understanding on the interplay between different assistance elements, and thus limited knowledge in how to choose between, or combine, them when designing AI systems. More problematically, studies are often driven by technical availability such as new explanation techniques. This practice may risk losing sight on what users need to better accomplish decision tasks or what are the necessary elements of the design space of human-AI decision making, which is especially important knowledge for practitioners to develop effective AI systems. For example, only a small number of studies explored non-model-centric system elements that can affect users' action space and cognitive processes, and showed that they are critical for user experience and also their interaction with model assistance features~\cite{buccinca2021trust,smithrenner2020, feng2018can}.

    \item \textbf{Focus on decision trials only instead of the holistic experience with AI } Existing work commonly experimented with participants performing discrete decision trial tasks---seeing an instance and making a decision with AI's assistance. However, in reality, when people use a decision-support AI system, many other steps and aspects can affect their experience and interaction with the AI, such as system on-boarding experience, workflow, contexts where the decision happens, and repeated experiences with the system. Their effects are currently under-explored for human-AI decision making. This narrow use of experimental tasks could have led to certain gaps or biases of assistance elements studied. For example, studies tended to focus on decision-specific assistance but less on model-wide information.
    
    \item \textbf{Gaps in models used.} Our analysis revealed a current bias in \textit{model types} used in the studies---more using traditional shallow models than deep learning models. It is necessary to elucidate how model types and their associated properties affect the experiment setup and generalizability of results, which can guide future studies to make appropriate choices. For example, some deep learning models not only tend to perform better in average (but not all) training settings, but also are likely less interpretable and prone to over-confidence. While wizard-of-oz approaches have a long tradition in HCI, applying them to studying AI models face many new challenges, such as how to simulate model errors and explanations in a realistic way. We caution against them without justifying the design as sufficiently approximating the interested model behaviors and stating the limitations. Another gap in models used is limited studies of \textit{regression} models. In addition to the prediction forms, some assistance elements take distinct forms for regression v.s. classification (e.g., uncertainty information), and their effects are under-explored for regression.
    
\end{enumerate}

\paragraph{Recommendations for future work.}

\begin{enumerate}

    \item \textbf{Human-centered analysis to define the design space of AI assistance elements.} Complementary to current practices of studying fragmented AI assistance elements, the field can benefit from having top-down frameworks that define the design space of AI assistance elements needed for better human-AI decision making, which requires analysis centering on what decision-makers need rather than technical availability.  Having this kind of framework can guide researchers to identify gaps in the literature and formulate research questions, and ultimately produce unified knowledge that can better help practitioners make appropriate design choices when developing AI decision support systems.  We hope our analysis can inform such efforts.
    \item \textbf{Extend the design space and studies beyond decision trials.} To center the research efforts on real user needs also means we should look beyond the discrete decision trials used by current studies, which not only lack may ecological validity but also fail to account for many temporal, contextual, and individual factors that can shape how people perceive and interact with AI, such as on-boarding experience, time constraints, workload, prior experience, and individual differences.  Future work should explore these factors, and conduct field and longitudinal studies of human-AI decision making.

   \item \textbf{Task-driven studies to complement feature-driven studies.} Current studies are often motivated by understanding the effect of certain assistance elements or design features. Then a decision task is chosen in an ad-hoc fashion or even arbitrarily in some cases. To inform the design space of AI assistance elements and actionable design guidelines for different types of AI system, we believe it is useful to complement current practices with task-driven studies, which may require conducting formative studies to understand user needs and behaviors for a given decision task.

\end{enumerate}

\section{Evaluation of Human-AI Decision Making}
\label{sec:metrics}

Deciding on the evaluation metrics is one of the most critical research design choices. This decision often involves choosing the construct --- \textit{what to evaluate}, then choosing the specific formulation or content of the metrics --- \textit{how} to evaluate the target construct. Our survey reveals a wide range of constructs evaluated in studies of human-AI decision making, likely due to broad research questions asked by the community and a lack of standardized evaluation methods. As mentioned in the Methodology section, our survey focuses on quantitative evaluation metrics, although some studies used qualitative analysis to gain a further understanding of user perceptions and behaviors.

At a high-level, we group the evaluation metrics into two categories:
(1) evaluation with respect to the decision making task and (2) evaluation with respect to the AI. Under each, we group them into areas of evaluation such as task efficacy versus efficiency. Then we further classify them as either objective or subjective measurements. Later in this section we discuss that subjective and objective measurements may in fact target different constructs (perception or attitude versus behavioral outcomes guided by the attitude). Here we classify them based on what the studies claim they are measuring. Note our analysis stays at the granularity of measurement construct instead of detailed differences in the content or formulation (e.g., what specific survey items are used). It is worth noting that many papers did not provide access to the survey scales or questionnaires. As a result, it can be difficult to interpret some of these findings or for future research to replicate them.

\subsection{Evaluation with respect to the decision task}

Decision making performance---for which in the AI assistance is designed to support---is intuitively the most important outcome measurement for human-AI decision making.
Evaluation of the decision tasks mainly belongs to two categories: (1) efficacy (i.e., the quality of the decisions);
and (2) efficiency (i.e., the speed in making these decisions).
In addition, we include a category on measuring people's task satisfaction.
Table~\ref{tb:metrics_task} gives a summary of these measures and how they are collected, both objectively and subjectively in the surveyed papers.

\begin{table}
\centering
\begin{tabular}{p{0.15\textwidth}|l|p{0.65\textwidth}}
\toprule
Evaluation & Subjective? & Metrics \\ \midrule
\multirow{2}{*}{Efficacy} & Subjective & Self-rated error/accuracy \citep{dietvorst2018overcoming,tsai2021exploring,lai2019human}, perceived performance improvement \citep{das2020leveraging}, confidence in the decisions \citep{green2019principles,logg2019algorithm,green2019disparate,guo2019visualizing}, 
soundness of participants' mental models \citep{kulesza2012tell}\\
& Objective & Accuracy/error~\citep{mcgrath2020does,lage2019evaluation, lucic2020does,liu2021understanding,bansal2019beyond,weerts2019human,jung2020limits,mallari2020look, zhang2020effect, lai2020chicago, lai2019human, cai2019human, gonzalez2020human,feng2018can,bansal2020does,buccinca2020proxy,buccinca2020proxy,hase2020evaluating,stowers2017insights,biran2017human,kocielnik2019will,ehrlich2011taking,lim2009and,kiani2020impact,nourani2021anchoring,grgic2019human,levy2021assessing,carton2020feature},
F1 \citep{narayanan2018humans,biran2017human}, precision \citep{biran2017human}, recall \citep{biran2017human,levy2021assessing}, AUC-ROC \citep{dressel2018accuracy}, 
false positive rate \citep{mallari2020look,green2019disparate,carton2020feature}, false negative rate \citep{dressel2018accuracy,carton2020feature}, true positive rate \citep{dressel2018accuracy},  true negative rate \citep{dressel2018accuracy}, 
Brier score \citep{green2019principles,green2019disparate},
 mean prediction error \citep{poursabzi2018manipulating}, 
win rate \citep{gero2020mental,das2020leveraging},  cumulative award \citep{gonzalez2020human}, customized score/return \citep{bansal2019updates}, 
 human percentile rank \citep{das2020leveraging}, 
 agreement between labels \citep{lee2021human}

\\
 \midrule
\multirow{2}{*}{Efficiency} 
& Objective & time taken on the task (response time, average time for a game round, speed) \citep{friedler2019assessing,lakkaraju2016interpretable,lage2019evaluation,narayanan2018humans,gero2020mental,weerts2019human,smithrenner2020,gonzalez2020human,yang2020visual,stowers2017insights,kocielnik2019will,arshad2015investigating,lim2009and,cheng2019explaining,abdul2020cogam,levy2021assessing,carton2020feature},
  total number of labels \citep{levy2021assessing}\\\midrule
\multirow{2}{0.15\textwidth}{Task satisfaction \& mental demand
} & Subjective & Satisfaction with the process \citep{dietvorst2018overcoming}, confidence in the process \citep{dietvorst2018overcoming}, frustration/annoyance \citep{kulesza2013too,smithrenner2020},  mental demand/effort \citep{kulesza2013too,weerts2019human,buccinca2020proxy,buccinca2021trust},
workload \citep{cai2019human,stowers2017insights,springer2018progressive,lee2021human}, 
task difficulty \citep{arshad2015investigating}
\\
& Objective & number of words in user feedback \citep{lakkaraju2016interpretable}\\
\bottomrule
\end{tabular}
\caption{Evaluation metrics with respect to the task. 
}
\label{tb:metrics_task}
\end{table}

\para{Efficacy.}
We start with objective metrics of decision task performance.
The most commonly used metric is \textit{accuracy}, measured as the percentage of correctly predicted instances (or equivalently, error rate, the percentage of incorrectly predicted instances)~\cite{bansal2019beyond,mallari2020look,dressel2018accuracy,ghai2020explainable,lakkaraju2016interpretable,nguyen2018comparing,lai2020chicago,lai2019human,feng2018can,chandrasekaran2018explanations,ribeiro2016should,lage2019evaluation,narayanan2018humans,lucic2020does,carton2020feature}.
Typically, the metric of interest is the accuracy of the joint outcome of human-AI teams, compared against the baseline accuracy of humans without AI assistance or of AI alone.

As the evaluation is essentially comparing decision labels against ground-truth labels,\footnote{As we discussed in \secref{sec:tasks}, the existence of ground-truth labels indicate that the task formulation tends to be objective in these studies.} other metrics that are typically used to evaluate AI performance can also be used to evaluate the performance of human-AI decision making.
These metrics include \textit{F1} \citep{narayanan2018humans,biran2017human}, \textit{precision} \citep{biran2017human}, \textit{recall} \citep{biran2017human,levy2021assessing}, \textit{AUC-ROC} \citep{dressel2018accuracy}, which are commonly adopted for imbalanced datasets in the machine learning literature.
For cases where the cost of mistakes varies significantly for the positive class and the negative class, \textit{false positives rate} \cite{mallari2020look,green2019disparate,dressel2018accuracy,carton2020feature} or
\textit{false negatives rate} \cite{mallari2020look,dressel2018accuracy,carton2020feature}, \textit{true positive rate} \citep{dressel2018accuracy}, and \textit{true negative rate} \citep{dressel2018accuracy} have been used. 
In regression tasks, such as asking people to predict the likelihood of recidivism \citet{green2019disparate,green2019principles},  prior work similarly adopts continuous counterparts of accuracy, including
\textit{mean prediction error}~\citep{poursabzi2018manipulating} and \textit{brier score} ($1 - (\text{prediction} - \text{outcome})^2$) \cite{green2019disparate,green2019principles}.

In gamified tasks, researchers also use \textit{win rate} \citep{gero2020mental,das2020leveraging},  \textit{cumulative award} \citep{gonzalez2020human}, \textit{customized return} \citep{bansal2019updates}, and \textit{human percentile rank} \citep{das2020leveraging} to capture the performance of human-AI teams.
Finally, 
in cases where groundtruth labels are not available, agreement between labels (\textit{inter-annotator agreement}) has also been used \citep{lee2021human}.

In addition to objective metrics, subjective metrics can help understand human perception of the task performance.
A natural extension to the objective metrics for performance is \textit{perceived accuracy} (i.e., self-rated error/accuracy) \citep{dietvorst2018overcoming,tsai2021exploring,lai2019human} and \textit{perceived performance improvement} \citep{das2020leveraging}.
Another common metric is to ask humans about their confidence in the decisions \citep{green2019principles,logg2019algorithm,green2019disparate,guo2019visualizing}.
These \textit{perceived confidence} measurements are usually based on Likert scales.

Finally, \citet{kulesza2012tell} introduce a unique metric that combines subjective metrics and objective metrics to measure \textit{mental model soundness} as, $\sum_i (\text{correctness}_i \times \text{confidence}_i)$, where $i$ is the index of questions. 
While this metric was originally used for comprehension questions of how a recommendation system works, \citet{kulesza2012tell} show it can be adapted to measure the soundness of mental models on test instances.

\para{Efficiency.}
In addition to efficacy---how \textit{accurately} participants make decisions, another important dimension to consider is efficiency---how \textit{quickly} they can make them.
The main motivation is to gauge if the AI assistance can help humans make decisions faster.
The most common objective metric is \textit{time taken on the task} \citep{friedler2019assessing,lakkaraju2016interpretable,lage2019evaluation,narayanan2018humans,gero2020mental,weerts2019human,smithrenner2020,gonzalez2020human,yang2020visual,stowers2017insights,kocielnik2019will,arshad2015investigating,lim2009and,cheng2019explaining,abdul2020cogam,levy2021assessing,carton2020feature}.
Alternatively, the \textit{total number of labels} (or \textit{task output}) can be used to measure efficiency as in \citet{levy2021assessing}, most appropriate when task time is held constant.
Notably, subjective metrics of efficiency are not seen in the papers we reviewed, although \textit{self-reported task efficiency} 
is a common metric used in usability testing~\cite{frokjaer2000measuring}.  

\para{Task-level satisfaction and mental demand.}
Finally, an important consideration in human-AI decision making is whether AI assistance improves human's satisfaction or enjoyment with the decision task. We consider both direct measurement of task satisfaction or the counter-measurement of task mental demand in this category. 

Most metrics in this category are subjective, typically solicited through survey questions at the exit survey or questionnaire. 
The only exception in papers we surveyed is \citet{lakkaraju2016interpretable}, who used the \textit{number of words in user feedback} to gauge user satisfaction.
Researchers have asked participants about their subjective \textit{satisfaction with the process} \citep{dietvorst2018overcoming}, \textit{confidence in the process} \citep{dietvorst2018overcoming}, \textit{frustration/annoyance} \citep{kulesza2013too,smithrenner2020}, \textit{mental demand/effort} \citep{kulesza2013too,weerts2019human,buccinca2020proxy,buccinca2021trust},
\textit{workload} \citep{cai2019human,stowers2017insights,springer2018progressive,lee2021human}, 
and \textit{task difficulty} \citep{arshad2015investigating}.

\subsection{Evaluation with respect to AI}
In addition to evaluating the decision task, works on human-AI decision making also focus on evaluating users' perception and response to the AI system itself, including understanding of AI, trust in AI, fairness perception, AI system satisfaction, and others. Table~\ref{tb:metrics_ai} summarizes these measures.

\begin{table}[]
\centering
\begin{tabular}{p{0.15\textwidth}|l|p{0.65\textwidth}}
\toprule
Evaluation & Subjective? & Metrics \\ \midrule
\multirow{2}{*}{Understanding} & Subjective &Self-reported understanding~\cite{buccinca2020proxy,anik2021data,yang2020visual,lucic2020does,smithrenner2020,binns2018s,wang2021explanations,cheng2019explaining,cai2019effects}, confidence in understanding~\cite{kulesza2012tell}, confidence in simulation~\cite{alqaraawi2020evaluating,nguyen2018comparing},  ease of understanding~\cite{guo2019visualizing,poursabzi2018manipulating}, intuitiveness~\cite{szymanski2021visual}, perceived transparency/interpretability~\cite{tsai2021exploring,rader2018explanations} \\
& Objective & Forward simulation~\cite{buccinca2020proxy,ribeiro2018anchors,wang2021explanations,yu2019trust,chromik2021think,nourani2021anchoring,alqaraawi2020evaluating,nguyen2018comparing,friedler2019assessing,chandrasekaran2018explanations,hase2020evaluating,poursabzi2018manipulating,abdul2020cogam}, counterfactual simulation~\cite{wang2021explanations,hase2020evaluating}, model errors detection~\cite{wang2021explanations}, identifying important features~\cite{szymanski2021visual,cheng2019explaining}, correctness of described model behaviors~\cite{rader2018explanations,kulesza2013too,chromik2021think}, correctness of estimated model performance/accuracy~\cite{smithrenner2020,dietvorst2015algorithm,green2019principles,nourani2021anchoring}, comprehension quiz~\cite{kulesza2012tell,gero2020mental,wang2021explanations,cheng2019explaining} \\ \midrule
\multirow{2}{*}{Trust and reliance} & Subjective & Self-reported trust~\cite{buccinca2020proxy,springer2018progressive,ribeiro2018anchors,tsai2021exploring,kulesza2012tell,smithrenner2020,gero2020mental,poursabzi2018manipulating,dietvorst2015algorithm,green2019principles,chromik2021think,alqaraawi2020evaluating,friedler2019assessing,poursabzi2018manipulating,cheng2019explaining,abdul2020cogam}, model confidence/acceptance~\cite{smithrenner2020,wang2021explanations,kulesza2013too,chromik2021think,szymanski2021visual,alqaraawi2020evaluating},  self-reported agreement/reliance~\cite{chandrasekaran2018explanations}, perceived accuracy~\cite{springer2018progressive,kocielnik2019will,smithrenner2020}, perceived capability/benevolence/integrity~\cite{rader2018explanations,nourani2021anchoring}, usage intention/willingness~\cite{kocielnik2019will,rader2018explanations,dietvorst2015algorithm,kulesza2013too,chromik2021think,nguyen2018comparing,abdul2020cogam,cai2019human} \\
& Objective & Agreement/acceptance of model suggestions~\cite{mcgrath2020does,bussone2015role,yin2019understanding,zhang2020effect,levy2021assessing,liu2021understanding,lai2019human,lai2020chicago,bansal2020does,wang2021explanations,lu2021human,yu2019trust,de2020case,biran2017human,carton2020feature}, switch~\cite{mcgrath2020does,yin2019understanding,zhang2020effect,park2019slow,lu2021human,grgic2019human}, weight of advice~\cite{poursabzi2018manipulating,logg2019algorithm},  model influence (difference between conditions)~\cite{green2019disparate,green2019principles},  disagreement/deviation~\cite{poursabzi2018manipulating}, choice to use the model~\cite{bansal2019beyond, dietvorst2018overcoming,ribeiro2016should,dietvorst2015algorithm}, over-reliance~\cite{yang2020visual,buccinca2021trust,bussone2015role,wang2021explanations}, under-reliance~\cite{yang2020visual,bussone2015role,wang2021explanations}, appropriate reliance~\cite{yang2020visual,poursabzi2018manipulating,wang2021explanations,gonzalez2020human}   \\\midrule
 \multirow{2}{0.15\textwidth}{Fairness} & Subjective & Perceived fairness~\cite{anik2021data,green2019disparate,harrison2020empirical,van2021effect,dodge2019explaining},  individual fairness~\cite{lee2019procedural}, group fairness~\cite{lee2019procedural}, process fairness~\cite{binns2018s}, deserved outcome~\cite{binns2018s}, feature fairness~\cite{binns2018s,van2021effect}, accountability~\cite{rader2018explanations} \\
& Objective & Decision bias~\cite{green2019principles,green2019disparate}  
 \\\midrule
 \multirow{2}{0.15\textwidth}{System satisfaction and usability} & Subjective & Satisfaction~\cite{biran2017human,tsai2021exploring,dietvorst2018overcoming,kocielnik2019will,lucic2020does,lage2019evaluation,narayanan2018humans}, helpfulness/support~\cite{ehrlich2011taking,buccinca2020proxy,yang2020visual,biran2017human,kocielnik2019will,cai2019human}, usefulness~\cite{guo2019visualizing,lee2020co}, effectiveness~\cite{tsai2021exploring}, quality~\cite{tsai2021exploring}, appropriateness~\cite{buccinca2020proxy}, preference/likability~\cite{yang2020visual,ribeiro2018anchors,kulesza2012tell,lee2020co}, system affect~\cite{springer2018progressive}, system usability~\cite{stowers2017insights}, complexity~\cite{buccinca2021trust}, ease/comfort of use~\cite{anik2021data,tsai2021exploring}, system frustration~\cite{kocielnik2019will,smithrenner2020,lee2020co},  richness/informativeness~\cite{anik2021data,lee2020co,lee2021human}, learning~\cite{tsai2021exploring}, recommendation to others~\cite{kocielnik2019will}  \\
& Objective & time spent on the application \citep{cai2019effects} 
 \\\midrule
{Others} & Subjective & Rate on specific features (e.g., explanations):  quality/soundness/completeness of explanation~\cite{szymanski2021visual,kulesza2013too,kunkel2019let}, usefulness/helpfulness of explanation~\cite{szymanski2021visual,nourani2021anchoring,cai2019human,lee2021human}, agreement with explanation~\cite{weerts2019human}, easiness to use explanation~\cite{szymanski2021visual}, explanation workload~\cite{abdul2020cogam}, attribution to AI versus self~\cite{cai2019effects}, desire to provide feedback~\cite{smithrenner2020}, expected model improvement~\cite{smithrenner2020}  \\
 
\bottomrule
\end{tabular}
\caption{Evaluation metrics with respect to AI.
}
\label{tb:metrics_ai}
\end{table}

\para{Understanding.} 
Since a significant proportion of empirical studies focus on AI explanations as a form of decision making assistance, users' understanding of the AI is a commonly used measurement.
Subjective metrics of understanding typically ask participants to directly rate their \textit{understanding} of the AI~\cite{buccinca2020proxy,anik2021data,yang2020visual,lucic2020does,smithrenner2020,binns2018s,wang2021explanations,cheng2019explaining,cai2019effects}, or some variations of it such as \textit{confidence in understanding}~\cite{kulesza2012tell}, \textit{ease of understanding}~\cite{guo2019visualizing,poursabzi2018manipulating}, or \textit{confidence in simulation}~\cite{alqaraawi2020evaluating,nguyen2018comparing}. Other metrics ask participants to rate on \textit{perceived intuitiveness}~\cite{szymanski2021visual} or \textit{transparency}~\cite{tsai2021exploring,rader2018explanations} of the AI system. 

Objective metrics often test how well people understand the system compared to ground-truth facts about its outputs or how it works.
The most commonly used metric is \textit{forward simulation}~\cite{buccinca2020proxy,ribeiro2018anchors,wang2021explanations,yu2019trust,chromik2021think,nourani2021anchoring,alqaraawi2020evaluating,nguyen2018comparing,friedler2019assessing,chandrasekaran2018explanations,hase2020evaluating,poursabzi2018manipulating,abdul2020cogam}, by asking participants to simulate a model's predictions on unseen instances. 
Some researchers have also used \textit{counterfactual simulation}~\cite{wang2021explanations,hase2020evaluating} (i.e., to predict feature changes that would lead to a different prediction). 
Other metrics measures the correctness of people's \textit{assessment of model performance}~\cite{smithrenner2020,dietvorst2015algorithm,green2019principles,nourani2021anchoring}, \textit{detection of errors}~\cite{wang2021explanations}, \textit{identification of important features}~\cite{szymanski2021visual,cheng2019explaining}, or whether they can provide a correct \textit{description of model behaviors}~\cite{rader2018explanations,kulesza2013too,chromik2021think}.
Other studies also design \textit{comprehension quizzes} to evaluate human understanding~\cite{kulesza2012tell,gero2020mental,wang2021explanations,cheng2019explaining}.
Using many of these objective measures, researchers aim to evaluate humans' \textit{mental model} of the AI systems' innerworkings and how they make predictions. 
It is important to note that objective and subjective understanding do not always align due to the phenomena of illusory confidence with which one \text{believes} they understood the model more than they actually did~\cite{chromik2021think}. 

\para{Trust and reliance.} Trust in AI is an important research topic and a commonly used metric. 
For subjective metrics, direct \textit{self-reported trust} 
is often used~\cite{buccinca2020proxy,springer2018progressive,ribeiro2018anchors,tsai2021exploring,kulesza2012tell,smithrenner2020,gero2020mental,poursabzi2018manipulating,dietvorst2015algorithm,green2019principles,chromik2021think,alqaraawi2020evaluating,friedler2019assessing,poursabzi2018manipulating,cheng2019explaining,abdul2020cogam} or some variations of it such as \textit{acceptance or confidence in the model}~\cite{smithrenner2020,wang2021explanations,kulesza2013too,chromik2021think,szymanski2021visual,alqaraawi2020evaluating}, \textit{self-reported agreement or reliance}
~\cite{chandrasekaran2018explanations}, or \textit{perceived accuracy} of the AI~\cite{springer2018progressive,kocielnik2019will,smithrenner2020}.  Some works measured user trust based on the well-established ABI framework~\cite{mayer1995integrative} or a subset of it, which measures the subjective trust belief (i.e., perceived trustworthiness) as \textit{perceived capability, benevolence and integrity}~\cite{rader2018explanations,nourani2021anchoring}, and/or \textit{trust intention as usage willingness}~\cite{kocielnik2019will,rader2018explanations,dietvorst2015algorithm,kulesza2013too,chromik2021think,nguyen2018comparing,abdul2020cogam,cai2019human}. 

Objective metrics of trust often focus on \textit{reliance} as a direct outcome of trusting (i.e., how much people's decisions rely on or are influenced by the AI's), such as \textit{acceptance of model suggestions}~\cite{bussone2015role,yin2019understanding,zhang2020effect,levy2021assessing,liu2021understanding,lai2019human,lai2020chicago,bansal2020does,wang2021explanations,lu2021human,yu2019trust,de2020case,biran2017human,mcgrath2020does,carton2020feature}, \textit{likelihood to switch}~\cite{yin2019understanding,zhang2020effect,park2019slow,lu2021human,grgic2019human,mcgrath2020does}, \textit{weight of model advice}~\cite{poursabzi2018manipulating,logg2019algorithm}, \textit{choice to use the model}~\cite{bansal2019beyond, dietvorst2018overcoming,ribeiro2016should,dietvorst2015algorithm}, model influence (difference between conditions)~\cite{green2019disparate,green2019principles},\footnote{This metric captures deviations from the control condition without model assistance.} as well as \textit{disagreement or deviation} from the model's recommendations~\cite{poursabzi2018manipulating}. 
Some researchers have also looked at more fine-grained reliance such as \textit{over-reliance} (relying on the model when it is wrong)~\cite{yang2020visual,buccinca2021trust,bussone2015role,wang2021explanations}, \textit{under-reliance} (not relying on the model when it is right)~\cite{yang2020visual,bussone2015role,wang2021explanations}, and \textit{appropriate reliance}~\cite{yang2020visual,poursabzi2018manipulating,wang2021explanations,gonzalez2020human}. 
These objective trust metrics often influence the joint decision-outcomes and thus correlate with the task efficacy metrics we reviewed above.

It is worth noting that although trust as an attitude guides the behavior of reliance, the two are in fact two different constructs. Social science and human factors literature have suggested that, besides trust, other factors can influence reliance, such as workload, time constraints, efforts to engage, perceived risk, and self-confidence (see a review in \citet{lee2004trust})

\para{Fairness.} Studying how people perceive the fairness of AI and what design impacts the perception is an active research area. These studies primarily rely on subjective metrics, from general \textit{perceived fairness}~\cite{anik2021data,green2019disparate,harrison2020empirical,van2021effect,dodge2019explaining} to perceptions of more fine-grained types of fairness such as \textit{individual fairness}~\cite{lee2019procedural}, \textit{group fairness}~\cite{lee2019procedural}, \textit{process fairness}~\cite{binns2018s}, \textit{deserved outcome}~\cite{binns2018s}, \textit{feature fairness}~\cite{binns2018s,van2021effect}, and \textit{accountability} (i.e., the extent to which participants think the system is fair and they can control the outputs the system produces)~\cite{rader2018explanations}. Only a small number of studies leveraged \textit{decision bias} (e.g., the action to follow model's recommendations despite their lack of fairness)~\citep{green2019principles,green2019disparate} as an objective metric of perceived fairness.

\para{System satisfaction and usability.} Many subjective metrics have been used to measure \textit{general satisfaction} with the AI~\cite{biran2017human,tsai2021exploring,dietvorst2018overcoming,kocielnik2019will,lucic2020does,lage2019evaluation,narayanan2018humans} or related constructs such as perceived \textit{helpfulness}~\cite{ehrlich2011taking,buccinca2020proxy,yang2020visual,biran2017human,kocielnik2019will,cai2019human}, \textit{usefulness}~\cite{guo2019visualizing,lee2020co}, \textit{effectiveness}~\cite{tsai2021exploring}, \textit{quality}~\cite{tsai2021exploring}, \textit{appropriateness}~\citep{buccinca2020proxy},
\textit{likeability}~\cite{yang2020visual,ribeiro2018anchors,kulesza2012tell,lee2020co}, etc. Some studies  leverage system usability related metrics such as \textit{usability}~\cite{stowers2017insights}, \textit{system complexity}~\cite{buccinca2021trust}, \textit{ease of use}~\cite{anik2021data,tsai2021exploring}, \textit{system frustration}~\cite{kocielnik2019will,smithrenner2020,lee2020co}, \textit{information richness}~\cite{anik2021data,lee2020co,lee2021human}, \textit{learning effect}~\cite{tsai2021exploring}, and \textit{recommendation to others} (net promoter)~\cite{kocielnik2019will}. 
 Of the papers we surveyed, only \citet{cai2019effects} leveraged objective satisfaction measures, using the \textit{time spent with the AI system} to reflect users' interest and satisfaction.

\para{Others.} Other measures focus on evaluating a specific feature of the AI. For example, AI explanations are frequently studied in the context of human-AI decision making, and subjective metrics have been used to measure \textit{people's perceived explanation quality}~\cite{szymanski2021visual,kulesza2013too,kunkel2019let}, \textit{explanation usefulness}~\cite{szymanski2021visual,nourani2021anchoring,cai2019human,lee2021human}, 
\textit{easiness to use explanation}~\cite{szymanski2021visual}, \textit{explanation workload}~\cite{abdul2020cogam},
and \textit{agreement with the explanation}~\cite{weerts2019human}.  
Other metrics include users' outcome \textit{attribution to AI versus self}~\cite{cai2019effects}, \textit{desire to provide feedback}~\cite{smithrenner2020}, and \textit{expected improvement of the AI system} over time~\cite{smithrenner2020}.

\para{Qualitative analysis.}
While the measurements described above are quantitative, we note that it is common for studies to supplement with qualitative analysis to further gauge the target measure (e.g., coded participants' statements about how the system work to measure understanding~\cite{kulesza2013too}) or understand the underlying mechanisms or reasons. For example, some studies asked open-ended ``why'' questions following survey scales, others conducted exit interviews or asked participants to think aloud while using the AI system~\cite{gero2020mental,binns2018s,veale2018fairness,kaur2020interpreting,bussone2015role,harrison2020empirical,cai2019hello,brown2019toward,smith2018closing,clark2018creative,smithrenner2020}.
Thematic analysis is then typically used to analyze these qualitative data, which allows researchers to extract main themes from the bulk of information and serve as insightful knowledge.
Other qualitative analysis performed includes grounded theory~\cite{liao2020questioning,muller2019data} and
affinity diagramming~\cite{beede2020human,yang2020re,yang2016investigating,holstein2019improving}.

\subsection{Summary \& Takeaways}
We summarize current trends and gaps in how surveyed studies evaluate human-AI decision making, and make recommendations for future work.

\paragraph{Current trends.}

\begin{enumerate}
    \item \textbf{Diverse evaluation focuses}. Depending on the research questions and assistance elements studies, prior studies focused on different evaluation constructs. Our analysis reveals a framework that differentiates between dimensions evaluating the human-AI decisions and dimensions evaluating human perception and interactions with the AI, each with subjective and objective measurements.
    \item \textbf{A focus on efficacy when evaluating decision tasks}, but efficiency and subjective satisfaction are also useful indicators.
    \item \textbf{Focuses on understanding, trust, system satisfaction and fairness with regard to AI.} We note that some of these focuses could be a result of the field's focus on explanation features and fair machine learning.

    \item \textbf{A lack of common measurements.} Within a given measurement area, there exists significant variations in the choices of evaluation construct, content, and formulation. For example, trust has been measured by a single item, multi-items, by trustworthiness dimensions, trust intention, and objective reliance, among others;  similarly, there are many nuanced constructs to measure satisfaction.
\end{enumerate}

\paragraph{Gaps in current practices.}
\begin{enumerate}

\item \textbf{A focus on decision efficacy (i.e., performance)}, and less emphasis on efficiency and user satisfaction. The three are commonly used constructs for usability measures~\cite{frokjaer2000measuring}.  This reflects a deep value of the field \citep{birhane2021values}, which prioritizes optimizing decision outcomes rather than the experience of human decision makers. That being said, we acknowledge that not all decision tasks require high efficiency  (also the efficiency of AI alone is trivially better than human-AI teams). An open question for the field is to better understand the role of efficiency in tasks where it is necessary for human and AI to collaborate.

\item \textbf{Use of subjective versus objective measurements need to be better understood and regulated.} It is important to recognize that the results from subjective and objective metrics do not always align, and the two may be in fact measuring different constructs, despite some studies make mixed claims. For example, participants can express high subjective understanding without objectively understanding the model behaviors. In some cases, objective metrics are measuring behavioral outcomes that are guided by user attitudes that are evaluated by subjective measures, but often in a non-linear way. One example that has long been studied in the human factors literature is trust as an attitude (by subjective measurements) versus reliance as a behavior (by objective measures). Despite some studies claim using reliance behaviors to reflect trust, many other factors besides trust can influence reliance, such as required efforts, perceived risk, self-confidence, and time constraints~\cite{lee2004trust}. We do not claim the superiority of either. Studies may choose to focus on objective versus subjective measurements for many reasons. For example, the research questions may deal with user attitude versus behavioral outcomes, or it is easier or only feasible for the system gather data for one type of measure in practice. However, this choice is often not explicitly justified or disentangled in terms of the actual constructs being measured.

\item \textbf{Home-grown measurements, especially subjective survey items, are often used.} There is a lack of practices to validate, re-use (and enabling re-use of) measurements, and leverage existing psychometrics or survey scales developed in HCI. Especially for subjective measurements, it is also not common practices to publish the survey scales used in the experiments. As a result, it can be difficult to replicate a study or compare different studies.

\item \textbf{Variance on the coverage of measurements.} Some studies measure only task efficacy or ask about user trust, other studies cover many aspects. While the choice should be driven by research questions, it might reflect a lack of common framework or awareness for researchers to make choices in a principled manner.

\end{enumerate}

\paragraph{Recommendations for future work.}

\begin{enumerate}

    \item \textbf{Make choices of evaluation metrics by research questions/hypotheses and targeted constructs. } It is important to articulate what constructs, whether it is subjective perception or attitude, objective behavioral outcomes, with regard to the AI or with regard to the decision tasks, should be measured for the research questions or hypotheses. In general, researchers should pay attention to the concepts of measurement validity established in statistics and social sciences, including construct validity (does the test measure the concept that it is intended to measure)~\cite{cronbach1955construct} and content validity (is the test fully representative of what it aims to measure)~\cite{lawshe1975quantitative}. Meanwhile, thinking through different areas of evaluation metrics (i.e., what can a given design/assistance element impact?) can help formulate more comprehensive and insightful hypotheses.

    \item \textbf{Work towards common metrics and a shared understanding on the meanings, strengths and weaknesses of different evaluation methods.} Such an understanding is key to a rigorous and replicable science. We must also recognize that human-AI decision making is a nascent area where new metrics may need to be developed. Studies should not be limited to focusing on areas reviewed in this paper or using existing metrics.
    
     \item \textbf{Keep reflecting on common evaluation metrics as value-laden choices.} Evaluation metrics, if widely accepted and used, can profoundly shape the outcomes of  a field. At a collective level, we should keep questioning whether the evaluation measures we use capture what matters for stakeholders and the society, and what the potential long-term outcome could be if we prioritize one set of measures over the other. This will also help the field expand the measurements and ultimately lead to more principled and responsible AI for decision making.
 
\end{enumerate}

\section{Summary: Towards a Science of Human-AI Decision Making}
\label{sec:discussion}

By summarizing decision choices made in more than 100 papers on empirical studies of human-AI decision making, specifically around the decision tasks, AI assistance elements studies, and evaluation metrics, we reflect on the barriers for the field to produce scientific knowledge and effectively advance human-AI decision making. A few core recommendations for future work emerged in our analysis, as we summarize below.

\para{Building on each other's work.} The advancement of empirical science requires joint effort. The field should learn from other experimental sciences such as psychology to practice replication, meta-analysis across studies, rigorous methodology and metrics development, and theory development that helps consolidate (sometimes contradicting) empirical results.  To build on each other's work also means that researchers should prioritize enabling others to re-use and reproduce when publishing results, by articulating rationales behind design choices and reflecting on them to build shared knowledge, and making study materials accessible. The field should also strive to establish common practices or infrastructure that make knowledge sharing easier. For example, in the context of evaluation for human-centered machine learning,  \citet{sperrle2021survey} propose the use of two artifacts: a checklist to help researchers make more principled choices in study design, and a reporting template that, besides results, covers many aspects of study design such as hypotheses, procedure, tasks, data, participants, and analysis.

\para{Developing common frameworks for human-AI decision making.}
Another aspect to enable generalizable and unified scientific knowledge is to develop frameworks that account for the research spaces for human-AI decision making. In this paper, we discuss the needs for the field to develop frameworks that characterize different decision tasks, lay out the design space for AI assistance elements, and areas of evaluation metrics. Such frameworks can help shape research efforts in several ways. First, they can provide researchers a shared understanding to identify important research problems and articulate research questions in a common language. For example, with a framework on the design space for AI assistance elements, researchers can identify under-explored areas. Second, frameworks make explicit otherwise latent or disregarded factors that can help interpret and consolidate results across studies, and ultimately lead to more robust knowledge and theories. For example, a framework on task characteristics can help differentiate between the setups of two studies, and a framework on evaluation metrics can help differentiate their coverage of measurements. Last but not least, developing principled frameworks is also a critical and reflective practice---reflecting on the limitations and gaps in current research, and questioning the missing perspectives. For example, we urge the field to consider the design space of AI assistance beyond supporting discrete decision trials by paying attention to the entire decision process, the holistic experience with an AI system, as well as contextual and individual factors. We also encourage research efforts that systematically examine what should be measured for human-AI decision making, considering what matters for different stakeholders instead of just the decision-makers (e.g., people whose life will be impacted by the decision), and what should be the ethical principles of decision-support AI.

\para{Bridging AI and HCI communities to mutually shape human-AI decision making.} To advance human-AI decision making requires both a foundational understanding of human needs and behaviors, and based on them developing more effective and human-compatible AI to support decision-makers. In the present time, the research efforts are somewhat one-directional. The HCI community typically work as the receiver of new AI techniques, then build systems or design evaluative studies. How can the two communities work better together? How can HCI research drive AI technical development? Such questions have been long contemplated on in other interdisciplinary areas such as interactive machine learning~\cite{amershi2014power} and human-robot interaction~\cite{vsabanovic2010robots}. We believe one aspect is to reconsider the priorities of HCI research contributions. Rather than focusing on conducting evaluative studies, developing theories and principled frameworks based on empirical studies and engagement with user needs can help guide AI research efforts. For example, a framework of AI assistance elements can inform what kinds of AI technique are needed to better support human-AI decision making; and a framework of evaluation metrics can guide the technical optimization efforts. Meanwhile, the AI communities should prioritize technical work that is informed by human needs and behaviors, and actively seek to distill insights from empirical studies as well as psychological and behavioral theories into computational work. As always, cross-disciplinary collaboration will require change of culture and translative research to bridge different perspectives. We hope the common goal of improving human-AI decision making can unite researchers from the two communities, and this survey as a bridge for joint research efforts.

\newpage
\bibliographystyle{ACM-Reference-Format}
\bibliography{refs}

\end{document}